\documentclass[10pt,twocolumn,letterpaper]{article}

\usepackage{cvpr}              

\usepackage{xspace}
\newcommand{\ours}{TerraScope\xspace}

\definecolor{cvprblue}{rgb}{0.21,0.49,0.74}
\definecolor{my_green}{RGB}{51,102,0}
\definecolor{my_red}{RGB}{204, 0, 0}
\usepackage[pagebackref,breaklinks,colorlinks,allcolors=cvprblue]{hyperref}
\usepackage{chngcntr}
\usepackage{graphicx}
\usepackage{multirow}
\usepackage{xcolor}
\usepackage{colortbl}
\usepackage{makecell} 
\usepackage{amssymb}
\usepackage{bbding}
\usepackage{pifont}
\usepackage{wasysym}
\usepackage{utfsym}
\usepackage{pifont}
\usepackage{fontawesome}
\usepackage{algorithmic}
\usepackage{algorithm}
\usepackage{stfloats}
\usepackage[most]{tcolorbox}

\definecolor{ModelGreen}{RGB}{213,232,212}

\def\paperID{*****} 
\def\confName{CVPR}
\def\confYear{2026}

\author{
Yan Shu\textsuperscript{1}\quad
Bin Ren\textsuperscript{1,4}\footnotemark[1] \quad
Zhitong Xiong\textsuperscript{3}\footnotemark[1] \quad 
Xiao Xiang Zhu\textsuperscript{3} \quad \\
Begüm Demir\textsuperscript{2} \quad
Nicu Sebe\textsuperscript{1} \quad
Paolo Rota\textsuperscript{1} \quad \\
[2mm]
\textsuperscript{1}~University of Trento \ \ 
\textsuperscript{2}~BIFOLD and TU Berlin \ \  
\textsuperscript{3}~Technical University of Munich \ \
\textsuperscript{4}~MBZUAI \ \
\\
[2mm]
\normalsize{
\url{https://shuyansy.github.io/terrascope/}
}
}

\title{\raisebox{-0.2cm}{\includegraphics[width=0.8cm]{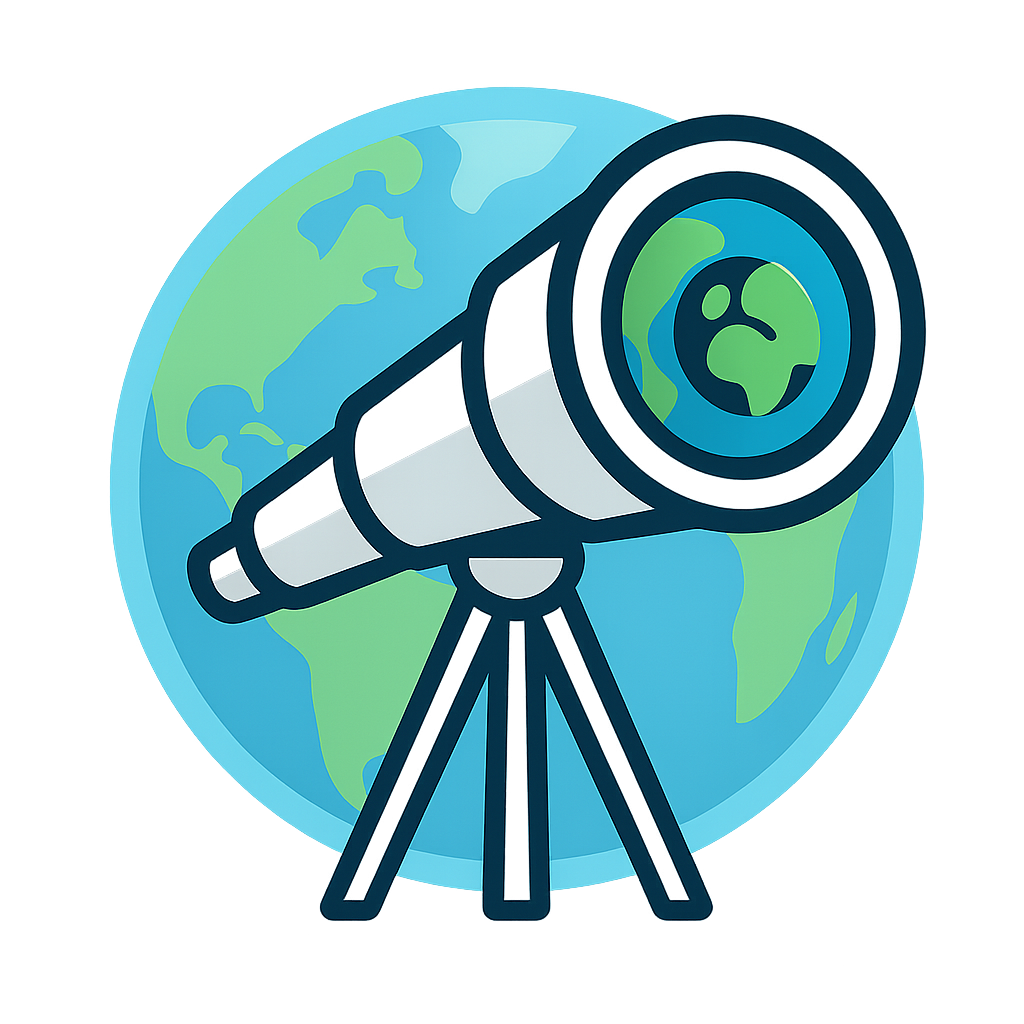}} TerraScope: Pixel-Grounded Visual Reasoning for Earth Observation}

\begin{document}


\twocolumn[{%
  \renewcommand\twocolumn[1][]{#1}%
  \maketitle
  \begin{center}
  \vspace{-9mm}
  \includegraphics[width=0.95\linewidth, trim=0 0 0pt 0, clip]{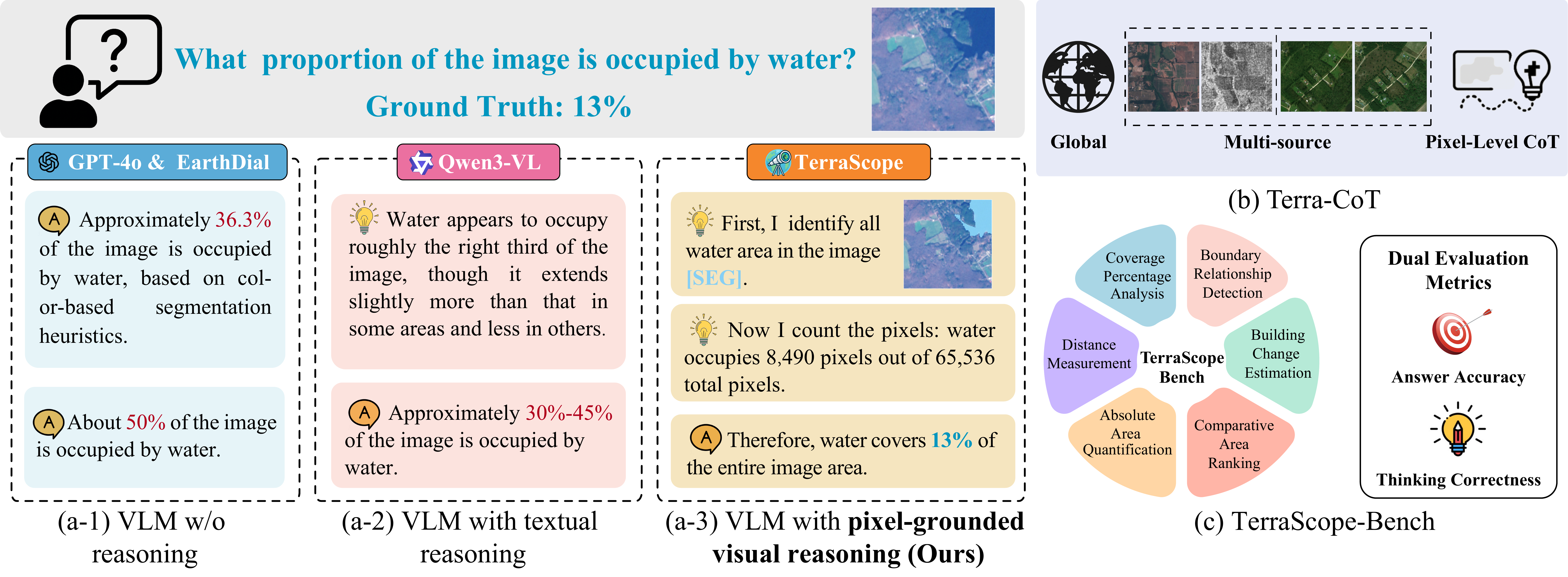}
  \end{center}
  \vspace{-6mm}
  \captionof{figure}{
      (a-1): The most common Vision Language Model (VLM) without reasoning directly outputs the wrong results.
      (a-2): Some solutions tried reasoning via textual Chain-of-Thought (CoT).
      (a-3): Our \ours, which takes the pixel-level grounding masks together with textual input, forming the interleaved CoT.
      (b): Our Terra-CoT 1M dataset.
      (c): Our TerraScope benchmark.
   }
   \label{fig:teaser}
   \vspace{4mm}
}]

\renewcommand{\thefootnote}{\fnsymbol{footnote}}
\footnotetext[1]{Correspondence to \texttt{<zhitong.xiong@tum.de>} and \texttt{<bin.ren@mbzuai.ac.ae>}}

\begin{abstract}
Vision-language models (VLMs) have shown promise in earth observation (EO), 
yet they struggle with tasks that require grounding complex spatial reasoning 
in precise pixel-level visual representations. To address this problem, we introduce TerraScope, a unified VLM that delivers pixel-grounded geospatial reasoning with two key capabilities: (1) \textit{modality-flexible reasoning}: it handles single-modality inputs (optical or SAR) and adaptively fuses different modalities into the reasoning 
process when both are available; (2) \textit{multi-temporal reasoning}: it 
integrates temporal sequences for change analysis across multiple time points. In addition, we curate Terra-CoT, a large-scale dataset containing 1 million 
samples with pixel-level masks embedded in reasoning chains across multiple 
sources.
We also propose TerraScope-Bench, the first benchmark for pixel-grounded 
geospatial reasoning with six sub-tasks that evaluates both answer accuracy 
and mask quality to ensure authentic pixel-grounded reasoning.
Experiments show that TerraScope significantly outperforms existing VLMs on pixel-grounded geospatial reasoning while providing interpretable visual evidence.
\end{abstract}    
\section{Introduction}
\label{sec:intro}
Earth observation (EO) satellites continuously monitor our planet at unprecedented scales, generating vast imagery archives for environmental monitoring \cite{xuan2025dynamicvl}, disaster response \cite{wang2025disasterm3}, and resource management \cite{kumar2022remote}. Traditional approaches \cite{sun2019remote,li2017integrating,li2025segearth,huang2025score} to EO data analysis rely on task-specific models, limiting flexibility across diverse applications. Vision-language models (VLMs) offer a paradigm shift: unified models that understand both visual content and natural language, enabling flexible analysis through text-based interaction. Recent domain-adapted VLMs have demonstrated strong performance on standard EO tasks, including image captioning \cite{lu2017exploring}, visual question answering \cite{lobry2020rsvqa,zhan2025skyeyegpt,luo2024skysensegpt,soni2024earthdial}, and visual grounding \cite{zhou2024geoground,zhan2023rsvg,zhang2024earthgpt,kuckreja2024geochat}, leveraging large-scale instruction tuning on remote sensing data.

However, state-of-the-art VLMs struggle with fine-grained geospatial 
reasoning requiring pixel-accurate spatial analysis. As illustrated in 
Fig.~\ref{fig:teaser}, leading general-purpose models (GPT-4o \cite{gpt4o}), 
reasoning-capable models (Qwen3-VL \cite{bai2025qwen2}), and EO-specific variants 
(EarthDial \cite{soni2024earthdial}) all fail to provide accurate answers 
on tasks such as calculating coverage of a land-cover class given in an image.
Recent multi-modal reasoning models \cite{zheng2025deepeyes,fan2025grit,zhang2025chain} have shown promise by grounding visual regions before reasoning. However, they cannot directly transfer to EO due to two fundamental differences:
(i) Unlike natural images with discrete objects, EO imagery depicts continuous spatial distributions where land cover types transition gradually. This continuous nature introduces substantial noise when coarse-grained grounding is used, hindering reasoning accuracy. 
(ii) EO analysis often involves multi-sensor and temporally evolving data. Optical imagery captures surface reflectance, SAR provides all-weather observation, and multi-temporal sequences reveal dynamic changes. However, existing VLMs struggle to effectively integrate such modality-flexible, time-varying data for EO reasoning within a single, unified framework.


To address these challenges, we present \ours, a comprehensive framework for pixel-grounded visual reasoning in EO. Building upon the recent paradigm of “\textit{thinking with images}”~\cite{su2025thinking}, \ours embodies the principle of “\textit{thinking with pixels}”: it explicitly localizes task-relevant regions and grounds each reasoning step in pixel-level visual evidence, rather than operating solely within the language domain.
Prior VLMs for EO rely on external tools~\cite{chen2024llm,shabbir2025thinkgeo,liu2024change,feng2025earth} for reasoning. The incorporation of external tools substantially increases the model’s complexity and reduces controllability, making it difficult to achieve pixel-level, intrinsic reasoning. In contrast, \ours employs mixed decoders that jointly generate segmentation masks and reasoning traces. The language model autonomously decides when to trigger mask generation and interleave the resulting visual tokens into the reasoning process, enabling dynamic visual grounding throughout multi-step reasoning. Beyond single-date single-modality data, \ours supports two independent reasoning capabilities. First, for \textit{multi-temporal reasoning}, it 
analyzes observations from multiple time points to deduce temporal changes 
based on evolving spatial patterns. Second, for \textit{multi-modal reasoning}, 
when both optical and SAR data are available, it adaptively selects the most 
informative modality for each reasoning step through text-guided cross-attention, 
leveraging optical for spectral information in clear regions while relying 
on SAR for cloud-covered areas.
To enable pixel-grounded reasoning at scale, we curate Terra-CoT, a 1M instruction-tuning dataset with pixel-level masks embedded in  reasoning traces generated via an automated pipeline, covering global scenes across multi-source EO data.
Additionally, existing EO benchmarks \cite{li2024vrsbench,wang2025xlrs,lobry2020rsvqa} primarily focus on visual perception tasks and lack evaluation of fine-grained visual reasoning capabilities. We introduce \ours-Bench, a benchmark specifically designed for pixel-grounded geospatial reasoning. It comprises 3,837 expert-verified questions supporting flexible evaluation with optical-only, SAR-only, or joint optical-SAR data, across both single-date and multi-temporal scenarios. Beyond traditional VQA accuracy metrics, \ours-Bench introduces dual evaluation metrics that assess both answer correctness and segmentation mask quality, ensuring models genuinely ground reasoning in pixel-level visual evidence.

In summary, our contributions are threefold:
\begin{itemize}

  \item We introduce \textbf{\ours}, a unified framework for pixel-grounded visual reasoning in EO. It grounds each reasoning step in precise segmentation masks for fine-grained, interpretable spatial analysis, supports multi-temporal change reasoning, and adaptively uses optical or SAR imagery.

  \item We curate \textbf{Terra-CoT}, a 1M instruction-tuning dataset with pixel-accurate masks embedded in reasoning traces, enabling scalable pixel-grounded training.

  \item We propose \textbf{\ours-Bench}, a benchmark of 3,837 expert-verified samples with dual metrics for answer accuracy and mask quality. Experiments on 11 models expose current limitations and demonstrate the effectiveness of \ours.

\end{itemize}
\begin{figure*}[!t]
    \vspace*{-1.5em} 
    \begin{center}
        \includegraphics[width=\textwidth]{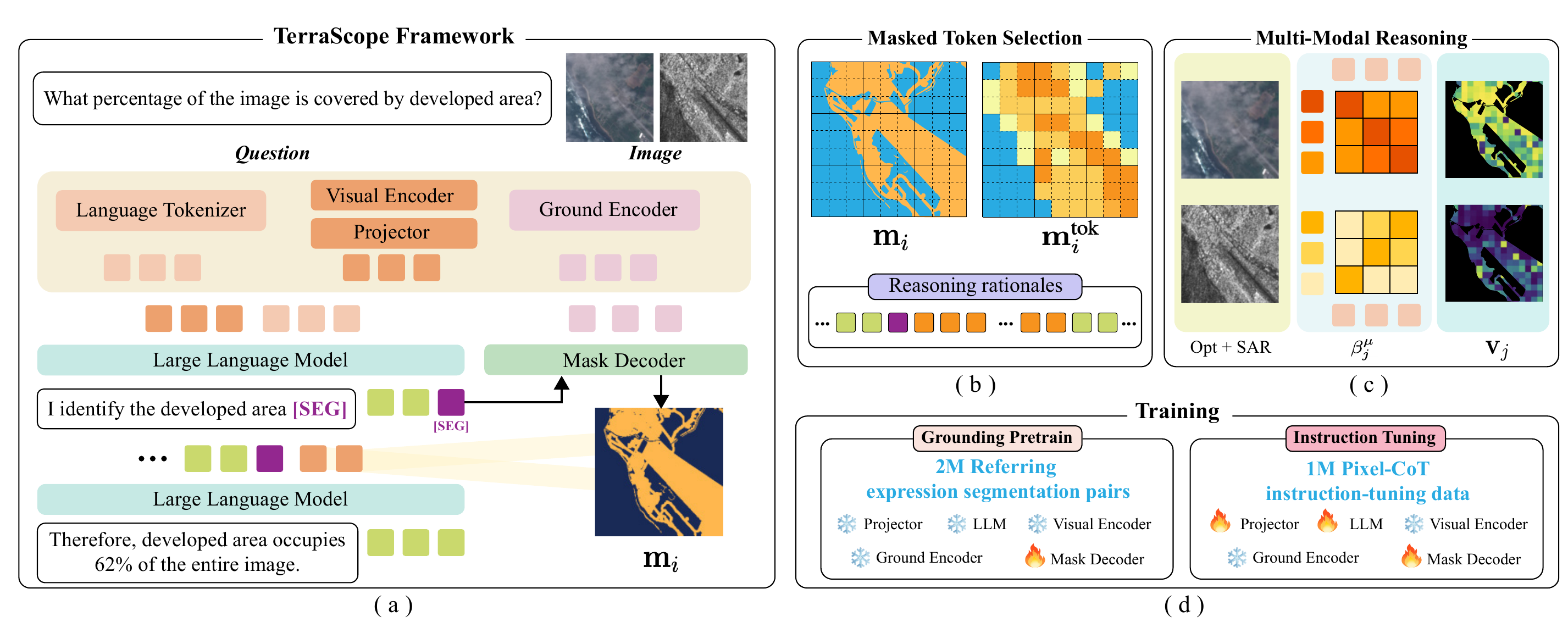}
    \end{center}
    \vspace{-20pt}
    \caption{
\textbf{Overview of TerraScope.} TerraScope generates textual reasoning 
tokens and segmentation masks in an interleaved manner, where masked visual 
features are injected at each reasoning step to ensure faithful pixel-grounded 
reasoning. TerraScope supports multi-modal and multi-temporal reasoning across EO data.}
 \label{fig:archi}
    \vspace{-10pt}
\end{figure*}

\section{Related Works}
\label{sec:relatedwork}
\noindent\textbf{Earth Observation VLMs.} Recent advancements in general-purpose 
VLMs \cite{li2024llava,liu2023visual,zhu2025internvl3} have shown 
impressive capabilities across various tasks. However, their limited 
exposure to remote sensing imagery hinders performance on EO
tasks. To address this gap, specialized EO-VLMs have emerged through domain-specific data curation and model adaptation. RSGPT \cite{hu2025rsgpt} enriches captioning 
datasets to enhance LLaVA's conversational abilities on satellite imagery. 
SkyEye-GPT \cite{zhan2025skyeyegpt} synthesizes 968K instruction samples for 
multi-task learning. Beyond image-level tasks, GeoChat \cite{kuckreja2024geochat}, 
SkySenseGPT \cite{luo2024skysensegpt}, and LHRS-Bot \cite{muhtar2024lhrs} 
incorporate visual grounding, region captioning, and reasoning. EarthGPT 
\cite{zhang2024earthgpt} introduces multi-sensor datasets spanning optical, 
SAR, and infrared modalities, while EarthMarker and EarthGPT-X \cite{zhang2025earthgpt} 
enable visual prompting interactions. GeoPixel \cite{shabbir2025geopixel} 
focuses on pixel-level grounding with grounded conversation datasets. EarthDial 
\cite{soni2024earthdial} scales multi-sensor data across multispectral, 
hyperspectral, and SAR to improve generalization. 
VHM \cite{pang2025vhm} 
proposes datasets with both factual and deceptive questions to improve model 
honesty. 
Despite these advances, existing EO-VLMs still lack pixel-grounded 
reasoning capabilities required for fine-grained spatial analysis.

\noindent\textbf{Earth Observation Benchmarks.} The rapid development of 
EO-VLMs has stimulated dedicated evaluation benchmarks. RSVQA \cite{lobry2020rsvqa}, 
LHRS-Bench \cite{muhtar2024lhrs}, RSIEval \cite{hu2025rsgpt}, and VLEO-Bench 
\cite{zhang2024good} evaluate conversational capabilities including classification, 
captioning, and VQA. VRSBench \cite{li2024vrsbench} and 
GeoChat-Bench \cite{kuckreja2024geochat} incorporate region-level grounding 
for localization evaluation. XLRS-Bench \cite{wang2025xlrs} focuses on 
ultra-high-resolution imagery understanding. GeoBench-VLM \cite{danish2024geobench} is a comprehensive benchmark covering multi-task and multi-sensor EO scenarios. DisasterM3 \cite{wang2025disasterm3} 
proposes a bi-temporal benchmark spanning multiple hazards, sensors, and tasks.
While recent benchmarks broaden the scope of sensors, tasks, and temporal settings, they still do not rigorously assess models’ capacity for pixel-accurate geospatial inference, leaving a gap in evaluating the precision needed for detailed spatial analysis.


\noindent\textbf{Visual Chain-of-Thought.} Recent works have explored 
grounding reasoning processes in visual content by interleaving visual 
evidence with textual reasoning chains. GRIT \cite{fan2025grit} interleaves 
the bounding box coordinates with the natural language reasoning for fine-grained 
counting. DeepEyes \cite{zheng2025deepeyes}, Chain-of-Focus \cite{zhang2025chain}, 
and Mini-o3 \cite{lai2025mini} employ iterative zoom-in mechanisms that crop 
and analyze focused regions. 
VLM-R1 \cite{shen2025vlm} and Visual-RFT 
\cite{liu2025visual} leverage reinforcement learning for visual grounding 
tasks. 
Mint-CoT \cite{chen2025mint} and ICoT \cite{gao2025interleaved} select 
relevant visual tokens through retrieval or attention mechanisms to compose 
multimodal rationales. 
However, these methods rely on coarse-grained spatial representations (bounding boxes, crops, or implicit token selection), which are inadequate for geospatial reasoning requiring pixel-level segmentation to capture continuous spatial distributions across multimodal data.












\begin{figure*}[!t]
    \vspace*{-1.5em} 
    \begin{center}
        \includegraphics[width=0.9\textwidth]{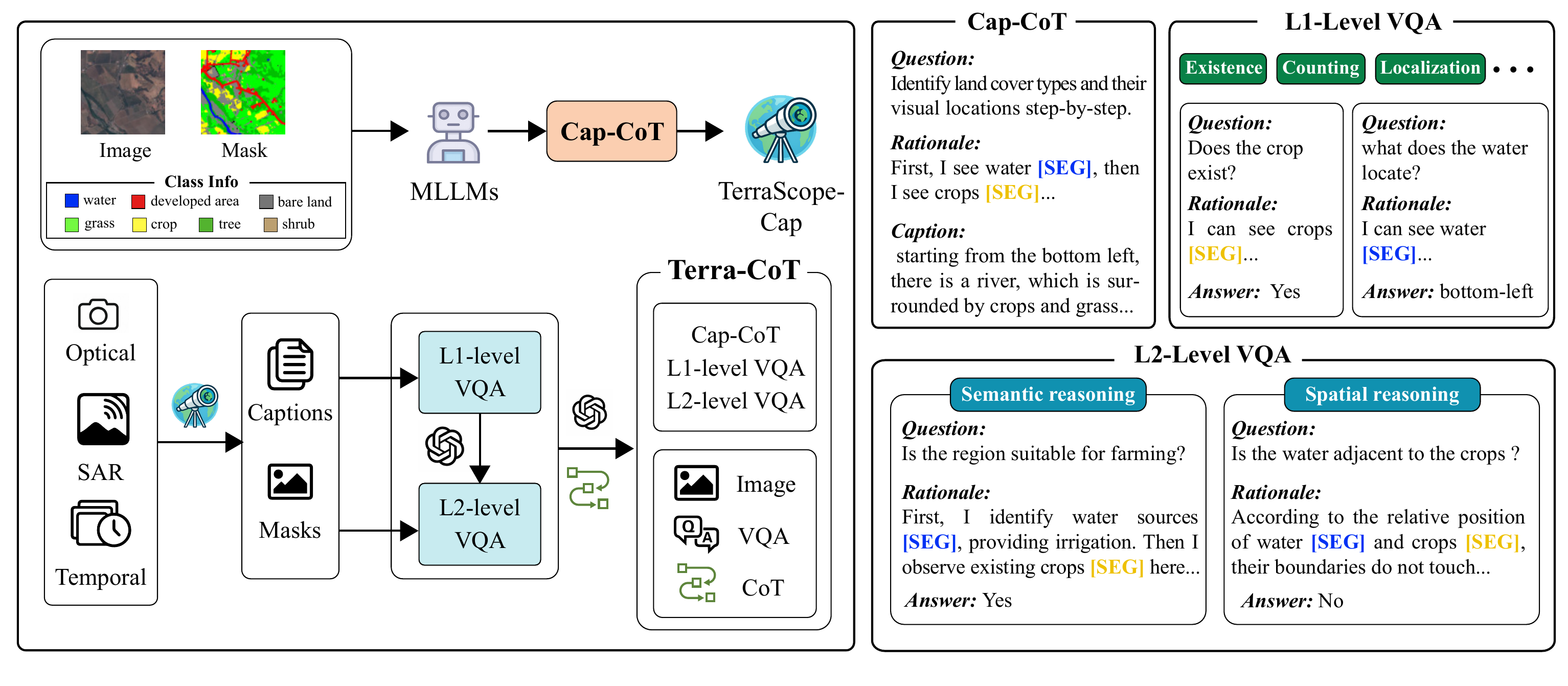}
    \end{center}
    \vspace{-25pt}
    \caption{
    \textbf{Terra-CoT curation pipeline.} 
    First, we generate Cap-CoT using ground truth masks and class labels to 
    train an initial annotation model. Second, we use the trained model to 
    annotate unlabeled data with pixel-accurate masks and captions. Third, based on the synthetic annotations, we apply hierarchical data synthesis to generate 
    diverse reasoning questions with chain-of-thought traces at two levels: 
    (L1) basic spatial grounding and (L2) complex multi-step reasoning including 
    spatial and semantic tasks.
    }
    \vspace{-15pt}
    \label{fig:terra-cot}
\end{figure*}

\section{Method}
\label{sec:method}


In this section, we present the core components of \ours and outline how pixel-grounded visual reasoning is formulated and implemented within our framework.

\subsection{Overview}

Geospatial reasoning demands fine-grained visual understanding that language-only 
reasoning cannot provide. In this context, we propose \textit{pixel-grounded 
visual reasoning}, where models explicitly generate segmentation masks and 
ground reasoning in the selected masked visual space. Formally, let $f(\cdot)$ 
be a VLM composed of a text encoder $f_T$ and a vision encoder $f_V$. Given 
a question $Q$ and an image $I$, the text encoder produces $\mathbf{q} = f_T(Q)$ 
and the vision encoder produces $\mathbf{v} = f_V(I) \in \mathbb{R}^{N \times D}$,where $N$ is the number of visual tokens and $D$ is the feature dimension.
Traditional VLMs then output an answer via language-only reasoning:

\begin{equation}
    [\mathbf{r}_1, \mathbf{r}_2, \ldots, \mathbf{r}_k, \mathbf{a}]  = f(\mathbf{v}, \mathbf{q}),
\end{equation}
where $k$ is the number of reasoning steps, $\mathbf{r}_i$ denotes the $i$-th 
textual reasoning step, and $\mathbf{a}$ is the final answer. Pixel-grounded visual reasoning interleaves masked visual features with textual reasoning:
\begin{equation}
    [\mathbf{r}_1, (\mathbf{m}_{1}, \mathbf{v}_{1}), \mathbf{r}_2, (\mathbf{m}_{2}, \mathbf{v}_{2}), \ldots, \mathbf{r}_k, (\mathbf{m}_{k}, \mathbf{v}_{k}), \mathbf{a}] = f(\mathbf{v}, \mathbf{q}),
\end{equation}
where at each reasoning step $i$, the model generates a segmentation mask 
$\mathbf{m}_{i}$ and selects masked visual features $\mathbf{v}_{i}$ from the 
identified regions.
In the rest of this section, we first present the TerraScope architecture that enables 
joint generation of masks and reasoning (Sec.~\ref{sec:TerraScope}), 
then describe Terra-CoT, our instruction dataset with interleaved visual 
and textual traces (Sec.~\ref{sec:Terra-CoT}).

\subsection{TerraScope Framework}
\label{sec:TerraScope}

As shown in Fig.~\ref{fig:archi}, our \ours builds upon a vision-language architecture augmented with a pixel-level segmentation module, forming a unified framework that integrates visual grounding and language-based reasoning within a single model.
Specifically, we leverage InternVL3 \cite{zhu2025internvl3} as our base model, which dynamically splits single images into sub-tiles while processing multi-image inputs independently, thereby defining a unified pipeline to transform all 
data into a uniform format. 

\noindent\textbf{Pixel-Grounded Chain-of-Thought.} The core innovation of TerraScope 
lies in the cooperative mechanism between dual decoders, which interleaves 
segmentation mask generation with text generation. Specifically, during the 
reasoning process, TerraScope monitors the language decoder's autoregressive 
output and triggers the mask decoder upon detecting \texttt{[SEG]}, which typically 
appears after mentions of key regions or objects. The mask decoder then predicts segmentation masks, from which masked visual 
tokens are selected and injected into the reasoning sequence to guide subsequent generation. For example, when answering ``Which is larger, water or road?'', 
the model generates ``I first identify water regions \texttt{[SEG]}...then 
road regions \texttt{[SEG]}'' and derives the answer by comparing their masked visual 
features.

As shown in Fig.~\ref{fig:archi} (b), to inject high-quality visual representations 
corresponding to the generated mask into reasoning traces, we first align 
the mask $\mathbf{m}_i$ with the visual encoder's dynamic patch layout by 
resizing it to the token grid resolution $(n \cdot s) \times (m \cdot s)$, 
where the image is split into $n \times m$ patches with each patch producing 
$s \times s$ tokens ($s=16$ for InternVL). To handle partial overlap between 
the pixel-level mask and token grid, we select a visual token if the mask 
covers more than 50\% of its corresponding spatial region. For the masked 
region, we extract the selected visual features as:
\begin{equation}
    \mathbf{v}_i = \{\mathbf{v}_j \mid \mathbf{m}_i^{\text{tok}}[j] = 1, j \in [1, N]\}
\end{equation}
where $\mathbf{v}_j$ denotes the $j$-th visual token in the feature map, and $\mathbf{m}_i^{\text{tok}}$ 
is the token-level mask derived from $\mathbf{m}_i$ by resizing to the token 
grid. The selected visual features $\mathbf{v}_i$ are then projected and 
flattened into a 1D sequence aligned with text embeddings, and fed into the 
LLM to resume autoregressive text generation conditioned on the KV cache of 
previously generated tokens.

\noindent\textbf{Multi-Modal and Temporal Reasoning.} Unlike single-image understanding, EO data often involves multi sources including optical-SAR pairs and temporal sequences. TerraScope handles these diverse scenarios through its flexible pixel-grounded reasoning framework.

For optical-SAR pairs, the model must identify complementary features, 
leveraging optical imagery for spectral information under clear conditions 
while relying on SAR for cloud-covered regions. We achieve this through 
text-guided, token-level modality selection. As shown in Fig.~\ref{fig:archi} (c), given optical and SAR images processed independently through the vision encoder 
to obtain visual features $\mathbf{v}_{\text{opt}}$ and $\mathbf{v}_{\text{SAR}}$, 
and question embeddings $\mathbf{q}$ from the text tokenizer with length $L$, 
we compute cross-attention between text and each visual modality, then aggregate 
across text tokens to obtain text-relevance scores:
\begin{equation}
    \beta^{\mu}_j = \frac{1}{L}\sum_{\ell=1}^{L} \text{Softmax}\left(\frac{\mathbf{v}^{\mu} \mathbf{q}^{\top}}{\sqrt{D}}\right)_{j\ell}, \quad \mu \in \{\text{opt}, \text{SAR}\}
\end{equation}
where $\beta^{\mu}_j$ denotes the relevance score of the $j$-th visual token 
to the question for modality $\mu$. 
When selecting masked visual features $\mathbf{v}_{i}$, we select features 
from the modality with a higher relevance score for each token position:
\begin{equation}
\small
    \mathbf{v}_j = \begin{cases}
        \mathbf{v}_j^{\text{opt}} & \text{if } \beta_j^{\text{opt}} > \beta_j^{\text{SAR}} \\
        \mathbf{v}_j^{\text{SAR}} & \text{otherwise}
    \end{cases}, \quad \forall j \text{ where } \mathbf{m}_i^{\text{tok}}[j] = 1
\end{equation}
This dynamic, spatially adaptive mechanism leverages the complementarity of paired EO data to boost reasoning.

For temporal sequences, a critical challenge is temporal disambiguation: 
when reasoning involves multiple observations, each \texttt{[SEG]} token 
must specify (1) which temporal image the mask decoder should segment from, 
and (2) from which image to extract the masked visual tokens. To address 
this, we incorporate explicit temporal indicators in the format ``Image: $t_i$'' 
before each \texttt{[SEG]} token. When the language decoder generates these 
signals, the mask decoder segments from image $t_i$ and the feature extraction 
module samples visual tokens from $\mathbf{v}^{(t_i)}$.
The model learns to generate timestamps from our Terra-CoT dataset, which contains temporally grounded reasoning traces paired with frame-specific masks (Sec.~\ref{sec:Terra-CoT}).

\noindent\textbf{Training.} We train TerraScope in two stages using supervised fine-tuning.  We first train on 2M referring expression 
segmentation pairs to establish basic grounding capability. We then fine-tune on 1M Terra-CoT samples to incentivize pixel-grounded visual reasoning ability. During training, we extract masked 
visual features from ground truth masks and interleave them into the sequence 
at positions following \texttt{[SEG]} tokens. The training objective combines language 
modeling loss (cross-entropy on text and \texttt{[SEG]} tokens, excluding injected 
visual features) and segmentation loss (Dice loss and pixel-wise cross-entropy):
\begin{equation}
    \mathcal{L} = \mathcal{L}_{\text{LM}} + \lambda \mathcal{L}_{\text{seg}},
\end{equation}
where we set $\lambda = 0.5$ to balance both objectives.

\subsection{Terra-CoT Dataset}
\label{sec:Terra-CoT}
Curating pixel-grounded visual CoT data is non-trivial: existing EO datasets provide either segmentation labels \cite{sumbul2019bigearthnet,yuan2024chatearthnet} or VQA pairs \cite{lobry2020rsvqa}, but not both with reasoning traces. We address this with a two-stage automated pipeline enabling large-scale pixel-grounded reasoning data.

\noindent\textbf{Grounded Captioning with Chain-of-Thought.} 
We leverage existing datasets with semantic annotations~\cite{yuan2024chatearthnet,sumbul2019bigearthnet,li2022mcanet} to construct pixel-grounded captioning data with reasoning traces (Cap-CoT). As shown in Fig.~\ref{fig:terra-cot}, we prompt a large multimodal model with an image where distinct land-cover categories are highlighted using colored masks and labeled accordingly. The model is instructed to produce detailed captions that explicitly reference these masked regions throughout its reasoning. 
This process yields 250K Cap-CoT samples, used to both train \ours and build an intermediate annotator, \ours-Cap, capable of generating pixel-grounded captions for unlabeled imagery.

\noindent\textbf{Hierarchical Data Synthesis.} 
Using \ours-Cap trained on Cap-CoT, we annotate images from diverse sources 
(optical, SAR, temporal) covering global regions with multi-category pixel-level 
labels (statistics in Appendix). Based on these annotations, we synthesize 
Terra-CoT through a two-level hierarchical process.

\textit{Level 1 (L1): Basic spatial grounding.} 
We generate template-based questions for randomly selected categories, covering 
fundamental spatial tasks such as existence verification, object counting, 
localization, area quantification, and boundary detection. For each question, 
we synthesize pixel-grounded reasoning traces using segmentation labels to 
explain the spatial analysis process.

\textit{Level 2 (L2): Complex multi-step reasoning.} 
We prompt an LLM to compose multiple L1 questions into complex reasoning tasks of two types:
(1) \textit{L2-Spatial} requires cross-entity 
spatial analysis such as relationship inference (\eg, ``Is the water adjacent 
to the crops?''); 
(2) \textit{L2-Semantic} requires domain knowledge beyond 
visual observation such as land suitability assessment (\eg, ``Is the region 
suitable for farming?''). 
For both types, the LLM synthesizes reasoning traces combining visual evidence with spatial or semantic analysis. This hierarchical process produces 1M Terra-CoT samples with diverse reasoning abilities.



\begin{figure}[!t]
    \vspace*{-1.5em} 
    \begin{center}
        \includegraphics[width=0.45\textwidth]{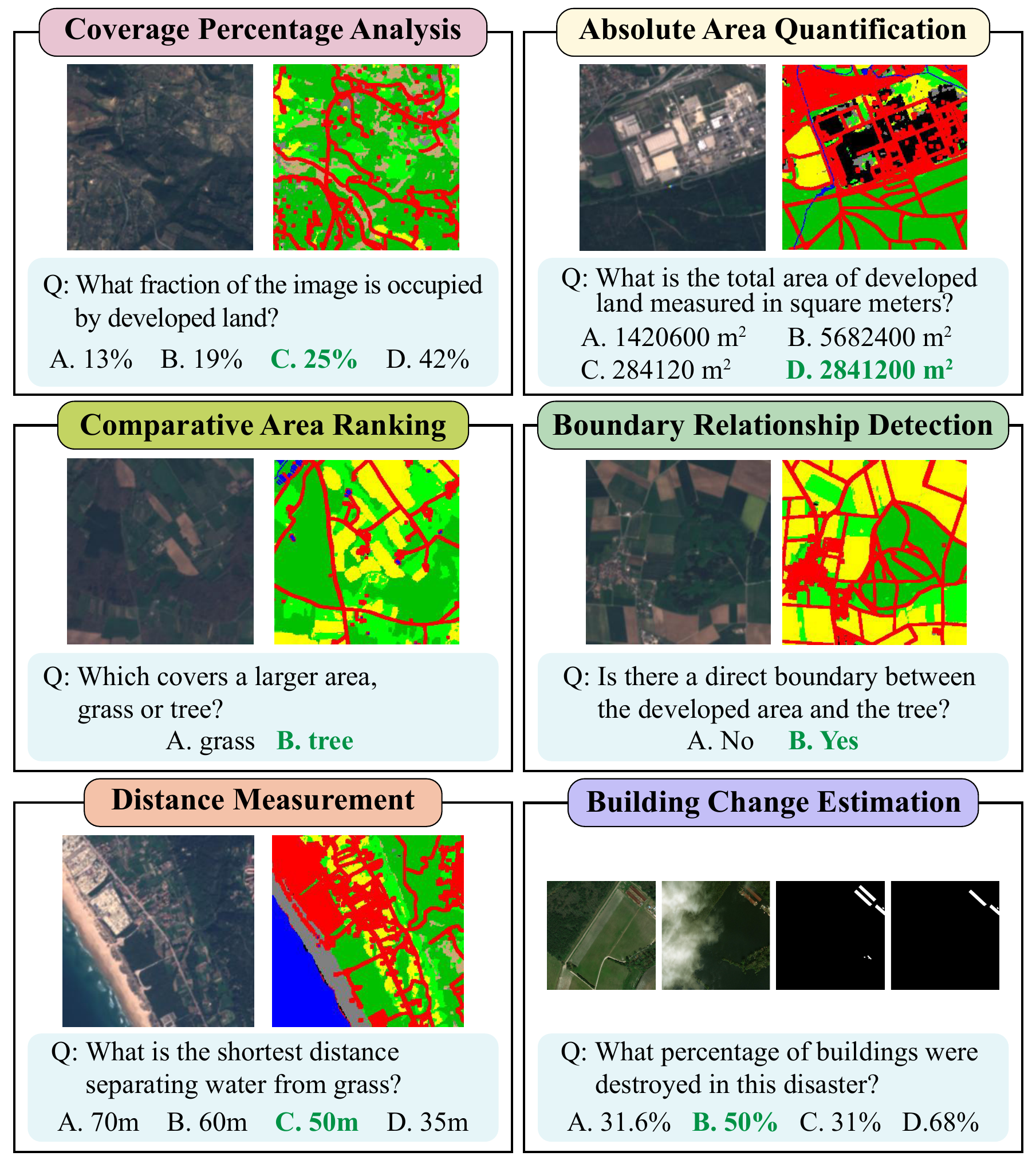}
    \end{center}
    \vspace{-20pt}
    \caption{Examples of TerraScope-Bench.}
    \vspace{-15pt}
    \label{fig:terrascopebench}
\end{figure}

\begin{table*}[!t]
    \scriptsize
    \centering
    \vspace{-4mm}
    \setlength\tabcolsep{7.2pt}
    \setlength{\extrarowheight}{0.1pt}
    \begin{tabular}{lc|ccccccc|cccc|ccc}
    \toprule
    \multicolumn{1}{c}{Model} & \multicolumn{1}{c|}{Size} 
    & \multicolumn{7}{c|}{\textbf{TerraScope-Bench}} 
    & \multicolumn{4}{c|}{\textbf{Landsat30AU}} 
    & \multicolumn{3}{c}{\textbf{DisasterM3}} \\
    \cmidrule(lr){3-9}\cmidrule(lr){10-13}\cmidrule(lr){14-16}
     & & CA & AQ & CR & BRD & DM & BCE & Avg. 
     & APR & NUM & SRI & Avg.
     & BDC & DRE & Avg. \\
    \midrule
    \rowcolor{gray!15}\multicolumn{16}{c}{\textbf{General VLMs}} \\
    GPT-4o$^{\dagger}$~\cite{gpt4o}            & - & 27.6 & 25.4 & 54.3 & 75.3 & 22.5 & 27.1 & 38.7 & - & - & - & - & 24.2 & 21.4 & 22.8 \\
    LLaVA-OV~\cite{li2024llava} & 7B & 28.0 & 21.2 & 56.6 & 75.9 & 19.4 & 23.7  & 37.5 & 39.4 & 46.6 & 85.1 & 57.0 & 26.4 & 24.2 & 25.3 \\
    Qwen2.5-VL~\cite{bai2025qwen2} & 7B &25.3 & 33.5 & 55.7 & 67.7 & 23.3 & 25.7 & 38.5 & 29.8 & 53.1 & \textbf{92.8} & 58.6 & 34.2 & 29.3 & 31.8 \\
    InternVL3~\cite{zhu2025internvl3} &8B & 22.3 & 26.3 & 57.2 & 67.0 & 18.6 & 24.3 & 36.0 & 31.4 & 42.4 & 90.6 & 54.8 & 30.3 & 24.1 & 27.2 \\
    GLM-4.1V-Think$^{\ddagger}$~\cite{hong2025glm} &9B & 24.8 & 57.1 & 55.2 & 58.4 & 23.3 & 29.5 & 41.4 & 45.7 & 58.6 & 70.0 & 58.1 & - & - & - \\
    Qwen3-VL-Think$^{\ddagger}$~\cite{bai2025qwen2} &8B & 29.0 & 47.8 & 57.9 & 67.8 & 25.6 & 31.9 & 43.3 &42.8 & 60.2 & 92.0 & 65.0 & 36.8 & 28.2 & 32.5 \\
    \midrule
    \rowcolor{gray!15}\multicolumn{16}{c}{\textbf{EO-Specific VLMs}} \\ 
    GeoChat~\cite{kuckreja2024geochat}    & 7B & 24.8 & 19.5 & 49.6 & 69.2 & 5.4 & - & 33.7 & 31.1 & 41.8 & 86.2 & 53.0 & - & - & - \\
    TeoChat~\cite{irvin2024teochat}          & 7B &25.6 & 17.8 & 55.8 & 55.8 & 8.5 & 22.6 & 31.0 & 30.2 & 59.6 & 87.1 & 59.0 & 22.5 & 23.3 & 22.9 \\
    LHRS-bot~\cite{muhtar2024lhrs}       & 7B & 13.7 & 24.3 & 54.0 & 28.4 & 12.4 & - & 26.6 & 63.5 & 12.5 & 82.6 & 52.9 & - & - & - \\
    EarthDial~\cite{soni2024earthdial}       & 4B & 26.3 & 24.1 & 54.4 & 69.2 & 20.2 & 23.6 & 36.3 & 23.5 & 43.6 & 51.2 & 39.4 & 30.2 & 20.8 & 25.5 \\
    EarthMind~\cite{shu2025earthmind}       & 4B &26.1 & 42.2 & 52.2 &73.3 & 38.1 & 20.8 & 42.1 & - & - & - & - & - & - & - \\
    \midrule
    \rowcolor{gray!15}\multicolumn{16}{c}{\textbf{Fine-tuned VLMs}} \\ 
    InternVL3~\cite{zhu2025internvl3}          & 8B &67.1 & 63.2 & 60.0 & 67.8 & 40.0 & 31.0 & 54.9 & 55.3 & 56.6 & 90.8 & 67.6 & 42.2 & 30.1 & 36.1 \\
    GLM-4.1V-Think$^{\ddagger}$~\cite{hong2025glm} &9B & 67.8 & 68.1 & 65.5 & 70.2 & 51.1 & 34.7 & 59.6 & 63.4 & 60.5 & 80.0 & 68.0 & 45.6 & 32.0 & 38.8 \\
    \midrule
    \rowcolor{ModelGreen}\textbf{TerraScope} & 8B & \textbf{73.2} & \textbf{70.2} & \textbf{71.8} & \textbf{80.0} & \textbf{65.9} & \textbf{52.1} & \textbf{68.9} & \textbf{69.8} & \textbf{60.8} & 91.1 & \textbf{73.9} & \textbf{54.1} & \textbf{38.9} & \textbf{46.5} \\ 
    \bottomrule
    \end{tabular}
    \vspace{-2mm}
    \caption{Quantitative performance of Terrascope on TerraScope-Bench (optical only), Landsat30AU and DisasterM3. ``Avg'' is the average performance of multiple-choice tasks. Bold means the best performance. $^{\dagger}$ denotes proprietary models. $^{\ddagger}$ denotes reasoning models. Fine-tuned VLMs are fine-tuned on our proposed Terra-CoT dataset.}
    \vspace{-5mm}
    \label{tab:eo_results_v3}
\end{table*}

\section{TerraScope-Bench}
EO imagery above 10m resolution presents unique challenges: 
individual objects span only a few pixels, and land-use boundaries become 
ambiguous, making precise pixel-level spatial reasoning essential. However, 
existing benchmarks (\eg, BigEarthNet \cite{sumbul2019bigearthnet}, ChatEarthNet \cite{yuan2024chatearthnet},) 
emphasize coarse-grained tasks such as scene classification and image captioning that depend primarily on global visual cues. As a result, they fail to adequately assess VLMs’ fine-grained reasoning capabilities, allowing models to perform well without genuine spatial understanding.

To address these limitations, we introduce \textbf{TerraScope-Bench}, a benchmark 
comprising 3,837 carefully curated samples from the test sets of existing datasets 
\cite{sumbul2019bigearthnet,yuan2024chatearthnet,gupta2019xbd}. As shown in 
Fig.~\ref{fig:terrascopebench}, our benchmark encompasses six task categories: Coverage Percentage Analysis (855), 
Absolute Area Quantification (855), 
Distance Measurement (129), 
Comparative Area Ranking (855), 
Boundary Relationship Detection (855), and Building Change Estimation (288).

We leverage pixel-level segmentation annotations to automatically generate 
question-answer pairs. For each sample, we compute spatial properties from 
segmentation masks, including coverage ratios, absolute areas, inter-object 
distances, and boundary relationships, to derive ground-truth answers. Questions 
are generated via templates to ensure diverse phrasing, then rephrased by an LLM 
to create natural variations and plausible distractors for multiple-choice format. 
Finally, human experts review the dataset to filter samples with erroneous masks. 
Unlike existing benchmarks that only assess final answer accuracy, TerraScope-Bench evaluates both response correctness and spatial reasoning quality using IoU-based segmentation metrics, verifying whether models attend to the correct regions during their reasoning process.

\begin{figure}[t]
    \begin{center}
        \includegraphics[width=0.4\textwidth]{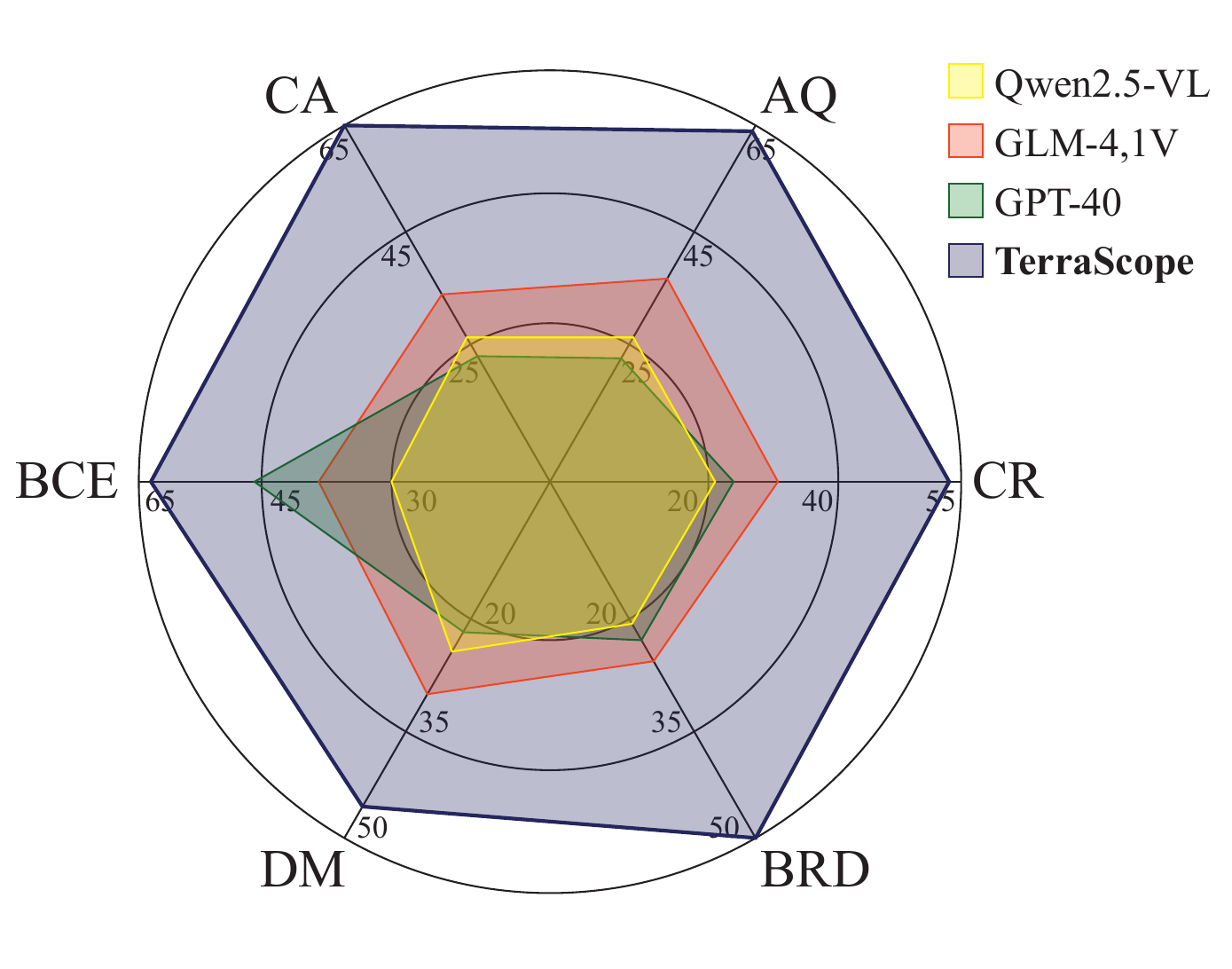}
    \end{center}
    \vspace{-30pt}
    \caption{Grounding IoU performance of different models.}
    \vspace{-20pt}
    \label{fig:ioutest}
\end{figure}

\section{Experiments}
\label{sec:experiments}

\noindent\textbf{Implementation details.}
Following the two-stage training strategy in Sec.~\ref{sec:TerraScope}, we first perform grounding pretraining where the vision encoder, projector, and LLM are frozen, and only the mask decoder is trained (lr=2e-5, batch size=8). In the second stage, we unfreeze the projector and mask decoder for full training and fine-tune the LLM via LoRA (lr=1e-5, batch size=2).
The vision encoder is kept frozen during training. All experiments run on NVIDIA H200-141GB GPUs, with additional dataset and hyperparameter details provided in the Appendix.

\noindent\textbf{Benchmarks.}
Beyond our proposed TerraScope-Bench, we evaluate TerraScope on two representative 
EO benchmarks in zero-shot settings to demonstrate its generalization 
capability. \textbf{LandSat30-AU} \cite{ma2025landsat30} features 30-meter 
resolution imagery with challenging reasoning subtasks; we report results on 
four tasks requiring fine-grained geospatial reasoning: Agro-Phenology Reasoning 
(APR), Numerosity Estimation (NUM), and 
Spatial-Relationship Inference (SRI). \textbf{DisasterM3} \cite{wang2025disasterm3} 
is a bi-temporal disaster assessment benchmark with pre- and post-event image 
pairs covering multi-hazard scenarios across multiple sensors; we evaluate on 
Damaged Building Counting (DBC) and Damaged Road Area Estimation (DRE).

\subsection{Main Results}
\label{sec:main_results}
We present the performance of TerraScope on several EO benchmarks 
in Tab.~\ref{tab:eo_results_v3}, where we evaluate 11 VLMs on TerraScope-Bench, 
including proprietary models and both general and EO-specific models. Additionally, 
we fine-tune InternVL3 and GLM-4.1V-Think on our Terra-CoT dataset to show 
its effectiveness. We highlight several key findings:

\noindent\textbf{(1) Pixel-grounded reasoning remains challenging.} Existing VLMs 
struggle with fine-grained geospatial reasoning, particularly on tasks requiring 
precise spatial analysis such as area percentage estimation. Both proprietary 
and open-source models achieve near-random performance, indicating 
the necessity of pixel-level grounding.

\noindent\textbf{(2) EO-specific models show limited advantages.} Despite training on 
large-scale EO data, EO-specific VLMs do not significantly outperform general 
VLMs on TerraScope-Bench. We hypothesize that this is because existing EO datasets 
predominantly feature high-resolution imagery ($<$5m), limiting models' ability 
to handle lower-resolution data prevalent in real-world applications.

    
\begin{table}[!t]
    \centering
    \footnotesize
    \setlength\tabcolsep{9pt}
    \setlength{\extrarowheight}{0.01pt}
    \begin{tabular}{>{\kern-0.5\tabcolsep}l|ccc<{\kern-0.5\tabcolsep}}
            \toprule
            \textbf{Model} & \textbf{TerraBen.} & \textbf{Landsat.} & \textbf{Disaster.}  \\
            \midrule
             Original  & 33.8 & 45.7 & 23.6 \\
            Textual CoT w/o Seg.  & 58.7 & 56.5 & 32.9 \\
            Textual CoT with Seg.   & 60.6 & 58.9 & 35.8 \\
            Random-Mask CoT  & 43.2 & 53.8 & 32.6 \\
            Box CoT  & 62.8 & 70.5 & 43.9 \\
            \midrule
            \rowcolor{ModelGreen}\textbf{TerraScope}  & \textbf{68.9} & \textbf{73.9} & \textbf{46.5}  \\
            \bottomrule
    \end{tabular}
    \vspace{-6pt}
    \caption{Ablation study on the effect of different CoT strategies for pixel-grounded visual reasoning. ``Original" denotes the base TerraScope model after pretraining, upon which we fine-tune with different CoT variations via SFT.}
    \vspace{-5mm}
    \label{tab:pgr}
\end{table}
\begin{figure}[!t]
    \centering
    \includegraphics[width=0.9\linewidth]{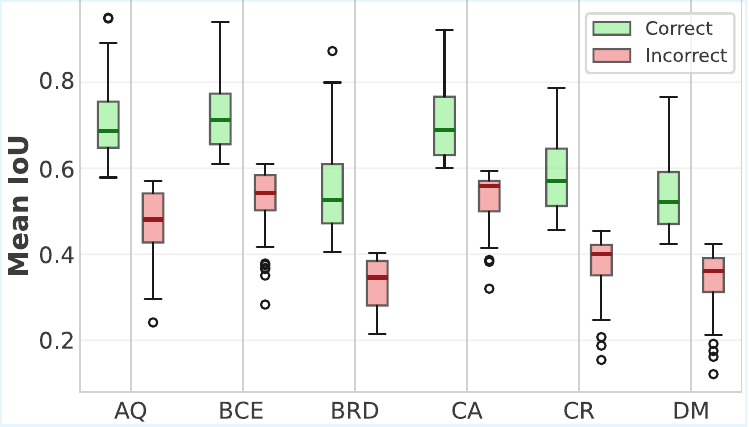}
    \vspace{-2mm}
    \caption{IoU distribution for correct vs. incorrect predictions.}
    \label{fig:person}
    \vspace{-6mm}
\end{figure}

\noindent\textbf{(3) Reasoning models perform better but lack visual grounding.} 
Models with explicit reasoning capabilities show stronger performance, especially on tasks requiring external knowledge like Absolute Area Quantification. However, their reasoning remains purely textual without grounding in pixel-level visual evidence, leading to hallucinations and insufficient fine-grained spatial perception.

\noindent\textbf{(4) Terra-CoT effectively improves model performance.} Fine-tuning 
general VLMs (\eg, InternVL3, GLM-4.1V-Think) on our Terra-CoT dataset leads to 
substantial performance gains across all tasks, demonstrating the effectiveness 
of our pixel-grounded reasoning data. However, challenging tasks like distance 
measurement (DM) and building change estimation (BCE) remain difficult, suggesting 
that data alone is insufficient and specialized architectural designs for 
pixel-grounded reasoning are necessary.

\noindent\textbf{(5) TerraScope achieves strong performance and generalization.} Our 
framework, which grounds reasoning in fine-grained visual perception, achieves 
the best results on TerraScope-Bench while demonstrating strong 
generalization to LandSat30-AU and DisasterM3.

\noindent\textbf{(6) TerraScope provides interpretable reasoning.} Beyond answer accuracy, 
TerraScope-Bench evaluates the reasoning process by measuring segmentation IoU 
against ground truth. As shown in Fig.~\ref{fig:ioutest}, TerraScope 
not only produces correct answers but also generates faithful reasoning traces 
with accurate spatial grounding, outperforming other grounding-capable models.

\subsection{Ablation Studies}
\label{sec:ablation}
We conduct extensive ablation studies to analyze TerraScope's effectiveness 
regarding its pixel-grounded visual reasoning mechanism and multi-modal reasoning, in which more details can be seen in Appendix.

\begin{table}[!t]
    \centering
    \footnotesize
    \setlength\tabcolsep{7pt}
    \setlength{\extrarowheight}{0.01pt}
    \begin{tabular}{>{\kern-0.5\tabcolsep}l|ccccc<{\kern-0.5\tabcolsep}}
        \toprule
        \textbf{Model} & \textbf{CA} & \textbf{AQ} & \textbf{CR} & \textbf{BRD} & \textbf{DM} \\
        \midrule
        No Fusion & 73.2 & 70.2 & 71.8 & 80.0 & 65.9 \\
        Concat. & \textbf{74.5} & \textbf{71.6} & \textbf{73.0} & \textbf{81.2} & 67.4 \\
        Text-guided (test only.) & 72.3 & 69.0 & 66.7 & 78.8 & 63.6 \\
        \rowcolor{ModelGreen}\textbf{Text-guided (train + test)} & 74.3 & 70.9 & 72.7 & 80.7 & \textbf{68.2} \\
        \bottomrule
    \end{tabular}
    \vspace{-6pt}
    \caption{Ablation study of multi-modal reasoning.}
    \label{tab:multimodal}
\end{table}

\begin{figure}[!t]
    \centering
    \vspace{-5mm}
    \includegraphics[width=0.85\linewidth]{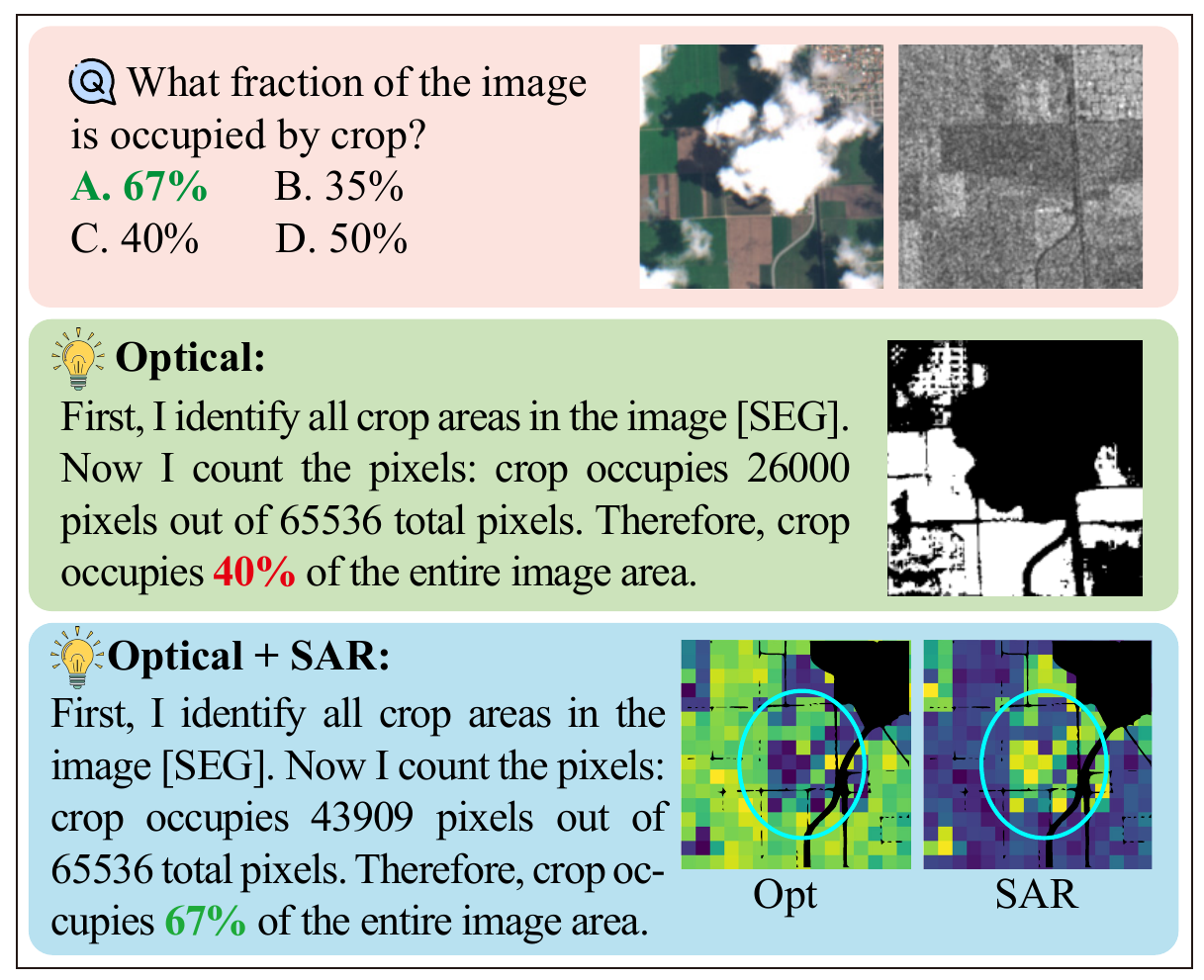}
    \vspace{-3mm}
    \caption{Visualizations of multi-modal reasoning.}
    \label{fig:modality_selection}
    \vspace{-6mm}
\end{figure}

\begin{figure*}[!t]
    \vspace*{-1.5em} 
    \begin{center}
        \includegraphics[width=0.9\textwidth]{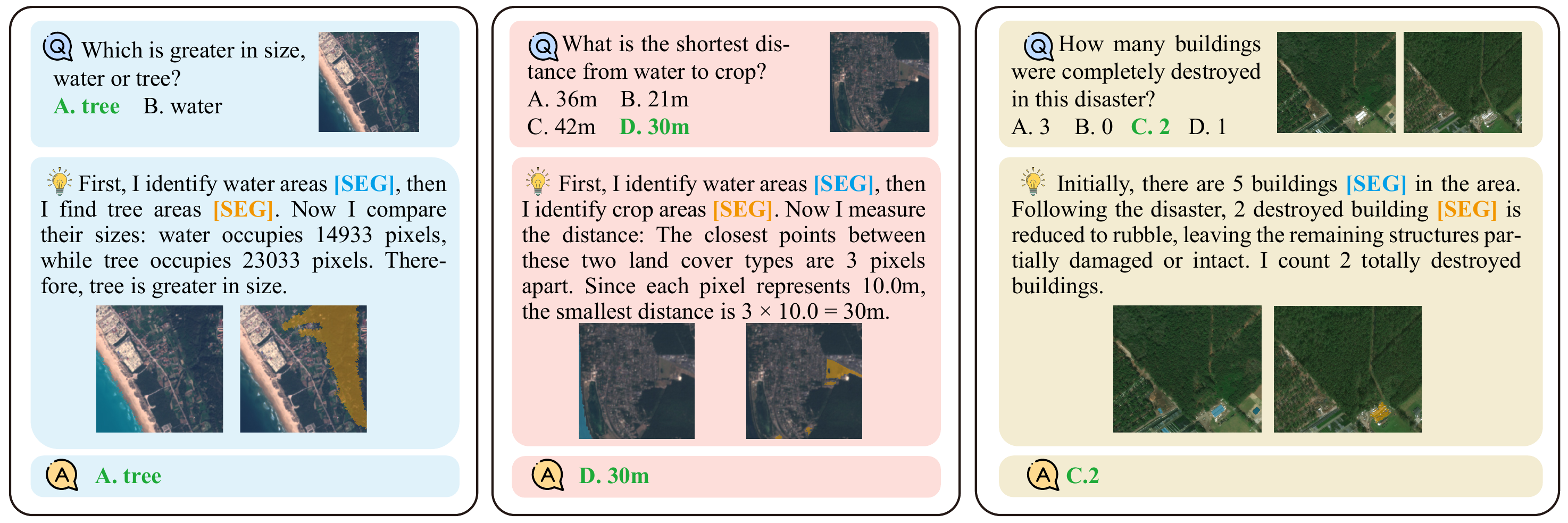}
    \end{center}
    \vspace{-20pt}
    \caption{Visualization of TerraScope.}
     \vspace{-6mm}
    \label{fig:viz}
\end{figure*}
\noindent\textbf{Effectiveness of pixel-grounded visual reasoning.} To verify that pixel-level grounding benefits reasoning, we compare several variants in Tab.~\ref{tab:pgr}. First, we train models with textual chain-of-thought 
only, where visual tokens are not interleaved into reasoning steps: either freezing the mask decoder (\textit{Textual CoT w/o Seg.}) or training it with ground truth masks as auxiliary supervision (\textit{Textual CoT w/ Seg.}). Results show that auxiliary segmentation training implicitly improves reasoning even when visual tokens are absent from the reasoning sequence, demonstrating the benefit of our joint training design.

Second, we examine the importance of mask-guided token selection. \textit{Random-Mask CoT}, which interleaves randomly selected visual tokens at each reasoning step 
without mask prediction, performs worse than textual CoT, likely due to irrelevant 
visual information hindering reasoning. \textit{Box CoT} uses the minimal 
bounding rectangles of predicted masks to select visual tokens, rather than 
using precise segmentation masks. This coarser grounding also underperforms 
TerraScope, especially on TerraScope-Bench and LandSat30-AU where land cover 
regions have irregular boundaries and shapes. These results confirm that precise 
pixel-level grounding through segmentation masks is essential for effective 
visual reasoning in EO.

Beyond final answer accuracy, we analyze the relationship between intermediate 
segmentation quality (measured by mean IoU with ground truth masks) and answer 
correctness. As shown in Fig.~\ref{fig:person}, samples with higher 
segmentation quality are significantly more likely to produce correct answers. 
Specifically, correct predictions achieve mean IoU of 0.628, substantially 
higher than incorrect predictions (0.443). The strong Pearson correlation ($r = 0.607$, $p < 0.001$) holds consistently across all task types ($r = 0.70$-$0.80$), demonstrating that  accurate pixel-level visual grounding is essential for correct geospatial reasoning.

\noindent\textbf{Effectiveness of multi-modal reasoning.} Beyond optical imagery, 
TerraScope can reason across optical and SAR modalities through text-guided 
modality selection. To validate its effectiveness, we compare several settings 
in Tab.~\ref{tab:multimodal}: (1) \textit{No fusion}: using only optical 
images; (2) \textit{Concat}: concatenating optical and SAR features and 
interleaving the concatenated features into each reasoning step; (3) \textit{Text-guided 
(test only)}: enabling modality selection only during inference; (4) \textit{Text-guided 
(train+test)}: enabling modality selection during both training and inference.

Results show that any form of multi-modal fusion substantially improves over 
the optical-only baseline. While \textit{Concat} achieves slightly higher 
accuracy than text-guided selection, our approach offers a critical advantage: 
it reduces context length by selecting only the relevant modality rather than 
processing both, improving efficiency while maintaining competitive performance. 
Importantly, training with modality selection is essential—enabling it only 
at test time yields no improvement, demonstrating that the model must learn 
when and how to leverage different modalities.

Fig.~\ref{fig:modality_selection} illustrates the effectiveness of our multi-modal reasoning through a cloud-contaminated case.
The figure 
reveals two key advantages: (1) \textbf{Improved segmentation through multi-modal 
fusion}: When using optical imagery alone, cloud cover causes severe segmentation 
errors in the masked regions. By fusing SAR data, which penetrates clouds, 
TerraScope produces accurate segmentation masks. (2) \textbf{Adaptive modality selection for reasoning}: 
Our text-guided selection mechanism adaptively chooses between optical and 
SAR based on data quality: it prioritizes optical tokens for cloud-free 
regions with reliable spectral information
while selecting SAR tokens 
for cloud-covered areas where optical data is corrupted. This dual advantage enables accurate reasoning under challenging conditions.


\section{Qualitative Results}
\label{sec:case_study}

Fig.~\ref{fig:viz} presents representative examples that demonstrating TerraScope's 
pixel-grounded reasoning capabilities across three challenging tasks: area 
percentage estimation, distance measurement, and temporal counting VQA. These 
cases illustrate TerraScope's dual strengths: (1) \textbf{Structured reasoning}: 
decomposing complex spatial questions into interpretable sub-steps through 
textual chain-of-thought, and (2) \textbf{Accurate visual grounding}: generating 
precise segmentation masks for relevant regions at each reasoning step. By 
grounding numerical computations in pixel-accurate visual evidence, TerraScope 
produces interpretable answers with transparent reasoning traces. Additional 
qualitative results and failure case analysis are provided in the Appendix.

\section{Conclusion}
\label{sec:conclusion}
In this paper, we presented \textit{\ours}, a unified vision-language framework 
for pixel-grounded geospatial reasoning in earth observation. By generating 
segmentation masks alongside reasoning traces, \ours achieves precise and 
interpretable spatial analysis, supporting multi-temporal 
change analysis and adaptive reasoning across optical and SAR modalities. 
We curated \textit{Terra-CoT}, a 1M instruction-tuning dataset with pixel-accurate 
masks embedded in reasoning chains, and introduced \textit{TerraScope-Bench}, the 
first benchmark for pixel-grounded geospatial reasoning. Extensive experiments validate the effectiveness of our approach across diverse geospatial reasoning tasks.

\clearpage
\section*{Acknowledgements}
This work was supported by the European Union Horizon projects ELIAS (No.\ 101120237) and ELLIOT (No.\ 101214398), and by the FIS project GUIDANCE (No.\ FIS2023-03251). Begüm Demir is supported by the European Research Council (ERC) through the ERC-2025-POC Agent-BigEarth Project under Grant 101292498. This work has been supported by Mountain Maps s.r.l.
{
    \small
    \bibliographystyle{ieeenat_fullname}
    \bibliography{main}
}

\clearpage
\setcounter{page}{1}
\maketitlesupplementary
\renewcommand{\thesection}{\Alph{section}}

\setcounter{section}{0}
\setcounter{figure}{0}    
\setcounter{table}{0}   

\renewcommand{\thetable}{\Alph{table}}
\renewcommand{\thefigure}{\Alph{figure}}
\renewcommand{\thesection}{\Alph{section}}



\section*{Appendix Overview}

\begin{itemize}
\item \ref{appendix:discussion}: \textbf{Limitations and Future work}.
\item \ref{appendix:comparison}: \textbf{Comparison to Concurrent Works}.
    \item \ref{appendix:architecture}: \textbf{Details of TerraScope}.
    \item  \ref{appendix:EarthMind-Bench}: \textbf{Details of TerraScope-Bench}.
      \item \ref{appendix:traindata}: \textbf{Details of Training Data}.
    \item  \ref{appendix:settings}: \textbf{Experimental Settings}.
    \item  \ref{appendix:efficiency}: \textbf{Efficiency Analysis}.
    \item \ref{appendix:ab}: \textbf{More Ablation Studies}.
    \item \ref{appendix:additional_results}: \textbf{Additional Experiment Results}.
     \item \ref{appendix:vis}: \textbf{Additional Visualizations and Failure Analysis}.
\end{itemize}

\section{Limitations and Future Work}
\label{appendix:discussion}
TerraScope focuses on pixel-level grounding for earth observation data, but
it has several limitations. First, like other multimodal large language models,
TerraScope may produce hallucinated outputs, \eg, generating plausible but
factually incorrect reasoning traces or inaccurate mask predictions that do
not correspond to actual ground features~\cite{bai2024hallucination,shu2025semantics}. Mitigating such hallucinations
through improved training strategies, verification mechanisms, or
retrieval-augmented generation is an important direction for future work.
Second, the interleaved generation of masks and reasoning traces increases
context length during training and inference. We analyze its computational
complexity in Sec.~\ref{appendix:efficiency}. A potential solution is to
compress masked visual tokens to reduce context length while retaining visual
grounding capability. Third, although TerraScope supports multi-sensor
reasoning, it currently handles only optical (RGB) and SAR data, with limited
capability for multi-spectral and hyper-spectral imagery~\cite{liu2025balanced}.
Future work will explore integrating these challenging data sources into the
reasoning framework. Finally, the current temporal reasoning capability of
TerraScope is limited to bi-temporal analysis (\ie, comparing two time points).
Many real-world EO applications, such as urban expansion monitoring,
deforestation tracking, and climate trend analysis, require reasoning over
long temporal sequences~\cite{shu2025video}. Extending TerraScope to support
multi-temporal and continuous time-series reasoning is an important direction
for future work.

\section{Comparison to Concurrent Works}
\label{appendix:comparison}
TerraScope belongs to the category of ``thinking with images'' models. In
Sec.~\ref{sec:intro} and Sec.~\ref{sec:relatedwork}, we described the
distinction between our approach and agent-based models. In this section,
we provide detailed comparisons with both unified interleaved reasoning models
and LLM-agent-based methods.

\noindent\textbf{Comparison with Unified Interleaved Reasoning Models.}
Several concurrent works share similar interleaved reasoning mechanisms with
TerraScope, including ICoT~\cite{gao2025interleaved}, GRIT~\cite{fan2025grit},
VGR~\cite{wang2025vgr}, and Mint-CoT~\cite{chen2025mint}. However, they differ
from TerraScope in two key aspects. First, these models are designed for
general vision tasks and have limited transferability to earth observation,
as they lack multi-modal reasoning (optical/SAR) and multi-temporal reasoning
capabilities essential for EO applications. Second, they employ different
mechanisms for interleaved reasoning:
\begin{itemize}[leftmargin=*]
\item \textbf{ICoT}~\cite{gao2025interleaved} proposes a training-free module
that leverages text-image cross-attention maps in LLMs to select relevant
tokens. However, this approach is limited to scenarios with salient objects
and fails when queries are complex or involve high-level semantic reasoning
not directly tied to visible objects.
\item \textbf{GRIT and VGR}~\cite{fan2025grit,wang2025vgr} use language to
model object coordinates (bounding boxes), which is inadequate for representing
pixel-level regions in EO data where spatial phenomena often lack clear
boundaries.
\item \textbf{Mint-CoT}~\cite{chen2025mint} overcomes bounding-box limitations
by selecting relevant image tokens through similarity-based implicit selection.
However, this approach may include tokens irrelevant to the current reasoning
step. To validate this, we trained Mint-CoT on our Terra-CoT dataset following
their official training paradigm, converting our pixel-level masks into their
token indices. Experiments (Tab.~\ref{tab:mintcot}) show Mint-CoT underperforms TerraScope on
TerraScope-Bench, confirming the importance of explicit mask generation for
pixel-grounded reasoning.
\end{itemize}

\noindent\textbf{Comparison with LLM-Agent-Based Methods.}
We further compare TerraScope with concurrent agentic approaches, including
ThinkGeo and EarthAgent. As shown in Tab.~\ref{tab:mintcot}, these methods
significantly underperform TerraScope. We attribute this to two main
limitations: (1)~\textit{Hallucination}: the LLM orchestrator may misinterpret
tool outputs or introduce reasoning errors during multi-step
planning~\cite{lin2025llm,li2026sponge}; (2)~\textit{Weak perception}: ThinkGeo
relies on box-level grounding, while EarthAgent adopts SAM-based grounding
with independently trained modules, limiting cross-module synergy. In contrast,
TerraScope's unified training paradigm enables bidirectional enhancement
between reasoning and pixel-level grounding, which agentic pipelines with
decoupled components cannot achieve.

\begin{table}[h]
\centering
    \vspace{-0.1in}
    \begin{tabular}{l|cc}
        \toprule
        \textbf{Methods}  & \textbf{TerraBench.} & \textbf{Landsat.}   \\
        \midrule
        \multicolumn{3}{l}{\textit{Interleaved Reasoning Models}} \\
         Mint-CoT (with SFT) &54.6 & 62.8   \\
         Mint-CoT (with SFT + RL)   &55.7 &63.2   \\
        \midrule
        \multicolumn{3}{l}{\textit{LLM-Agent-Based Methods}} \\
         ThinkGeo & 28.5 & --   \\
         EarthAgent & 37.6 & --   \\
        \midrule
        \rowcolor{ModelGreen}\textbf{TerraScope} & \textbf{68.9} & \textbf{73.9}    \\
        \bottomrule
    \end{tabular}
    \caption{Comparison of TerraScope with interleaved reasoning models and LLM-agent-based methods on TerraScope-Bench.}
    \label{tab:mintcot}
\end{table}

\begin{algorithm}[t]
\small
\caption{TerraScope Inference}
\begin{algorithmic}[1]
    \STATE \textbf{Input:} Question embeddings $\mathbf{q}$, Visual features $\mathbf{v}$ (or $\mathbf{v}^{\text{opt}}$, $\mathbf{v}^{\text{SAR}}$), Mask decoder $f_{\text{mask}}$, Max tokens $\lambda$, Stopping criteria $SC$
    \STATE \textbf{Output:} Generated answer $\mathbf{a}$ with reasoning traces
    \STATE $\mathrm{predicted\_tokens} \gets []$ \textcolor{blue}{\COMMENT{Initialize as empty list}}
    \STATE $\mathrm{reasoning\_step} \gets 0$ \textcolor{blue}{\COMMENT{Track reasoning step index}}
    \STATE $\mathrm{inputs} \gets$ Initialize($\mathbf{q}$, $\mathbf{v}$) \textcolor{blue}{\COMMENT{Initialize inputs for prefilling}}
    
    \textcolor{blue}{\COMMENT{Compute modality relevance scores if multi-modal}}
    \IF{both $\mathbf{v}^{\text{opt}}$ and $\mathbf{v}^{\text{SAR}}$ are available}
        \STATE $\beta_j^{\mu} \gets \frac{1}{L}\sum_{\ell=1}^{L} \text{Softmax}\left(\frac{\mathbf{v}^{\mu} \mathbf{q}^{\top}}{\sqrt{D}}\right)_{j\ell}$ for $\mu \in \{\text{opt}, \text{SAR}\}$
    \ENDIF
    
    \WHILE{$SC$ not met}
        \STATE $\mathrm{next\_token}, \mathrm{hidden\_state} \gets \textbf{LLM}(\mathrm{inputs})$
        \STATE Append $\mathrm{next\_token}$ to $\mathrm{predicted\_tokens}$
        
        \textcolor{blue}{\COMMENT{Check if \texttt{[SEG]} token is generated}}
        \IF{$\mathrm{next\_token} = \texttt{[SEG]}$} 
            \STATE $\mathrm{reasoning\_step} \gets \mathrm{reasoning\_step} + 1$
            \STATE $i \gets \mathrm{reasoning\_step}$
            
            \textcolor{blue}{\COMMENT{Generate segmentation mask}}
            \STATE $\mathbf{m}_i \gets f_{\text{mask}}(\mathrm{hidden\_state})$ 
            
            \textcolor{blue}{\COMMENT{Resize mask to token grid}}
            \STATE $\mathbf{m}_i^{\text{tok}} \gets$ ResizeToTokenGrid($\mathbf{m}_i$)
            
            \textcolor{blue}{\COMMENT{Select tokens with $>50\%$ coverage}}
            \STATE $\mathcal{I} \gets \{j \mid \text{Coverage}(\mathbf{m}_i^{\text{tok}}, j) > 0.5\}$
            
            \textcolor{blue}{\COMMENT{Apply spatial sampling if exceeds threshold}}
            \IF{$|\mathcal{I}| > \lambda$}
                \STATE $\mathcal{I} \gets$ SpatialUniformSample($\mathcal{I}$, $\mathbf{m}_i^{\text{tok}}$, $\lambda$)
            \ENDIF
            
            \textcolor{blue}{\COMMENT{Extract masked visual features}}
            \IF{both modalities available}
                \FOR{$j \in \mathcal{I}$}
                    \IF{$\beta_j^{\text{opt}} > \beta_j^{\text{SAR}}$}
                        \STATE $\mathbf{v}_j \gets \mathbf{v}_j^{\text{opt}}$
                    \ELSE
                        \STATE $\mathbf{v}_j \gets \mathbf{v}_j^{\text{SAR}}$
                    \ENDIF
                \ENDFOR
            \ELSE
                \STATE Extract features from single modality
            \ENDIF
            
            \STATE $\mathbf{v}_i \gets \{\mathbf{v}_j \mid j \in \mathcal{I}\}$
            \STATE Append $\mathbf{v}_i$ to $\mathrm{predicted\_tokens}$
            
        \ENDIF 
        
        \STATE $\mathrm{inputs} \gets$ Update($\mathrm{inputs}$, $\mathrm{predicted\_tokens}$) 
        \textcolor{blue}{\COMMENT{Update KV cache for next generation}}
    \ENDWHILE
    
    \STATE $\mathbf{a} \gets$ Tokenizer.decode($\mathrm{predicted\_tokens}$)
    \STATE \textbf{return} $\mathbf{a}$
\end{algorithmic}
\label{alg:terrascope_inference}
\end{algorithm}

\section{Details of TerraScope}
\label{appendix:architecture}

\noindent\textbf{Vision-Language Model.}
The VLM component of TerraScope is built upon InternVL-3~\cite{zhu2025internvl3}. 
In InternVL-3, each image is divided into multiple patches at a pre-defined 
resolution ($448 \times 448$). Each patch is processed by the vision encoder 
and encoded into 256 tokens. For instance, an image with 4 patches (plus one 
global thumbnail) yields $(4+1) \times 256 = 1{,}280$ visual tokens in total. For multi-temporal inputs, we do not split images into patches 
but directly feed independent images into the model. For example, for a multi-temporal sequence with 
$T$ observations, the total number of visual tokens is $T \times 256$.

\noindent\textbf{Pixel-Grounding Module.}
TerraScope's pixel-grounding module is initialized with the pre-trained 
SAM-2 model~\cite{ravi2024sam}. We connect SAM-2 and the LLM via the special 
token \texttt{[SEG]}. The hidden states of the \texttt{[SEG]} token from the last layer of LLM serve 
as a spatial prompt and are fed into SAM-2's decoder, which generates 
segmentation masks. This design allows the LLM to control mask generation 
through learned prompt embeddings.

During training, the SAM-2 decoder is fine-tuned to understand the spatial 
prompts, and gradients are backpropagated through the \texttt{[SEG]} token 
to the LLM, enabling it to generate better prompts. During inference, if 
the LLM does not generate a \texttt{[SEG]} token, we interpret this as 
indicating that no segmentation is needed for the current reasoning step.

\noindent\textbf{Masked Token Selection.}
To balance effectiveness and efficiency, we set a maximum threshold $\lambda = 128$ 
for the number of visual tokens in $\mathbf{v}_i$. If the number of selected 
tokens exceeds this threshold, we apply spatial uniform sampling to retain 
$\lambda$ tokens while preserving spatial coverage. Specifically, we divide 
the masked region into a $\lceil\sqrt{\lambda}\rceil \times \lceil\sqrt{\lambda}\rceil$ 
grid and select one token from each grid cell, choosing the token closest 
to the cell center. This ensures representative spatial sampling across the 
entire masked region rather than biased concentration in any local area.

\noindent\textbf{Inference Process.}
TerraScope performs autoregressive generation with pixel-grounded reasoning 
(Algorithm~\ref{alg:terrascope_inference}). The vision encoder processes 
input images to obtain visual features $\mathbf{v}$ (or $\mathbf{v}^{\text{opt}}$, 
$\mathbf{v}^{\text{SAR}}$ for multi-modal inputs), which are cached for 
efficiency. At each step, the LLM generates the next token. When a \texttt{[SEG]} 
token is generated, TerraScope: (1) generates a segmentation mask $\mathbf{m}_i$ 
via the mask decoder conditioned on the \texttt{[SEG]} token's hidden states; 
(2) extracts masked visual features $\mathbf{v}_i$ by selecting tokens with 
$>50\%$ coverage and applying spatial uniform sampling if the count exceeds 
$\lambda=128$; (3) for multi-modal inputs, adaptively selects between optical 
and SAR based on text-relevance scores $\beta_j^{\mu}$. The selected features 
$\mathbf{v}_i$ are then injected into the generation sequence, and the LLM 
continues reasoning conditioned on both textual and visual contexts through 
KV cache updates.

\begin{figure*}[!t]
    \vspace*{-1.5em} 
    \begin{center}
        \includegraphics[width=0.9\textwidth]{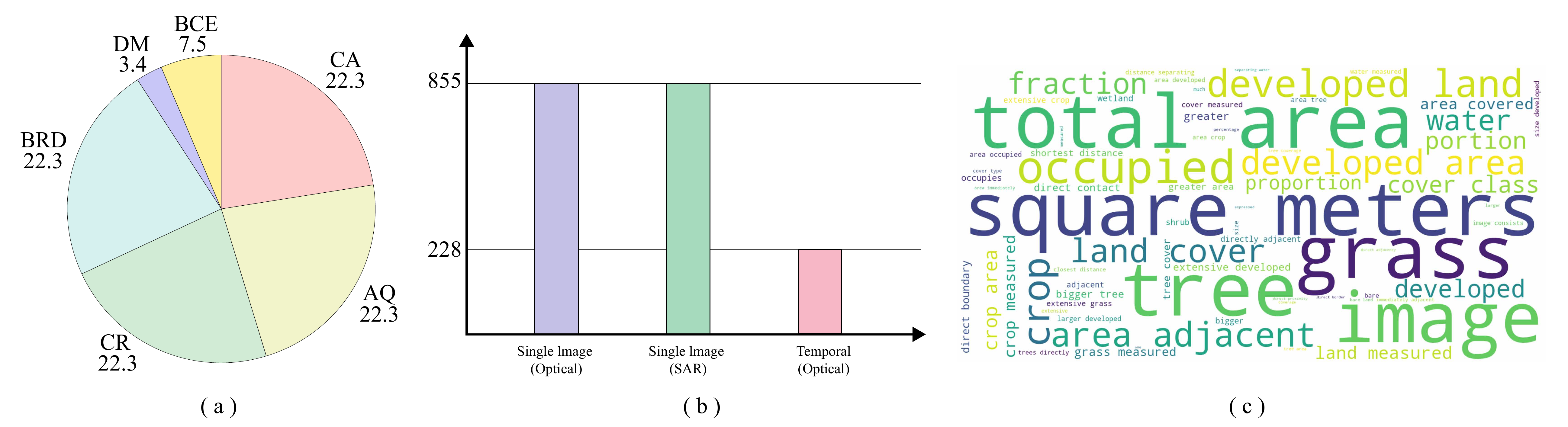}
    \end{center}
    \vspace{-20pt}
    \caption{Data distribution of TerraScope-Bench.}
     \vspace{-2mm}
    \label{fig:terrabench_sta}
\end{figure*}

\section{Details of TerraScope-Bench}
\label{appendix:EarthMind-Bench}

\subsection{Overview}
We present a more detailed analysis of TerraScope-Bench in Fig.~\ref{fig:terrabench_sta}. Subfigures (a–c) illustrate the distribution of task categories, image source (multi-sensor and multi-temporal) and the visualization of word clouds of question, showing that TerraScope-Bench covers a wide variety of object types and semantics, enabling comprehensive evaluation across pixel-level grounded visual reasoning tasks.

\subsection{Data Annotations for TerraScope-Bench}
\label{appendix:data_annotations}

TerraScope-Bench consists of six task types requiring pixel-grounded reasoning. 
We construct the benchmark through a three-stage pipeline: (1) heuristic-based 
answer generation from pixel-level annotations, (2) GPT-4o-based question 
rephrasing and distractor generation, and (3) expert validation and quality 
control.

\begin{figure*}[h]
    \vspace*{-2mm} 
    \begin{center}
        \includegraphics[width=0.9\textwidth]{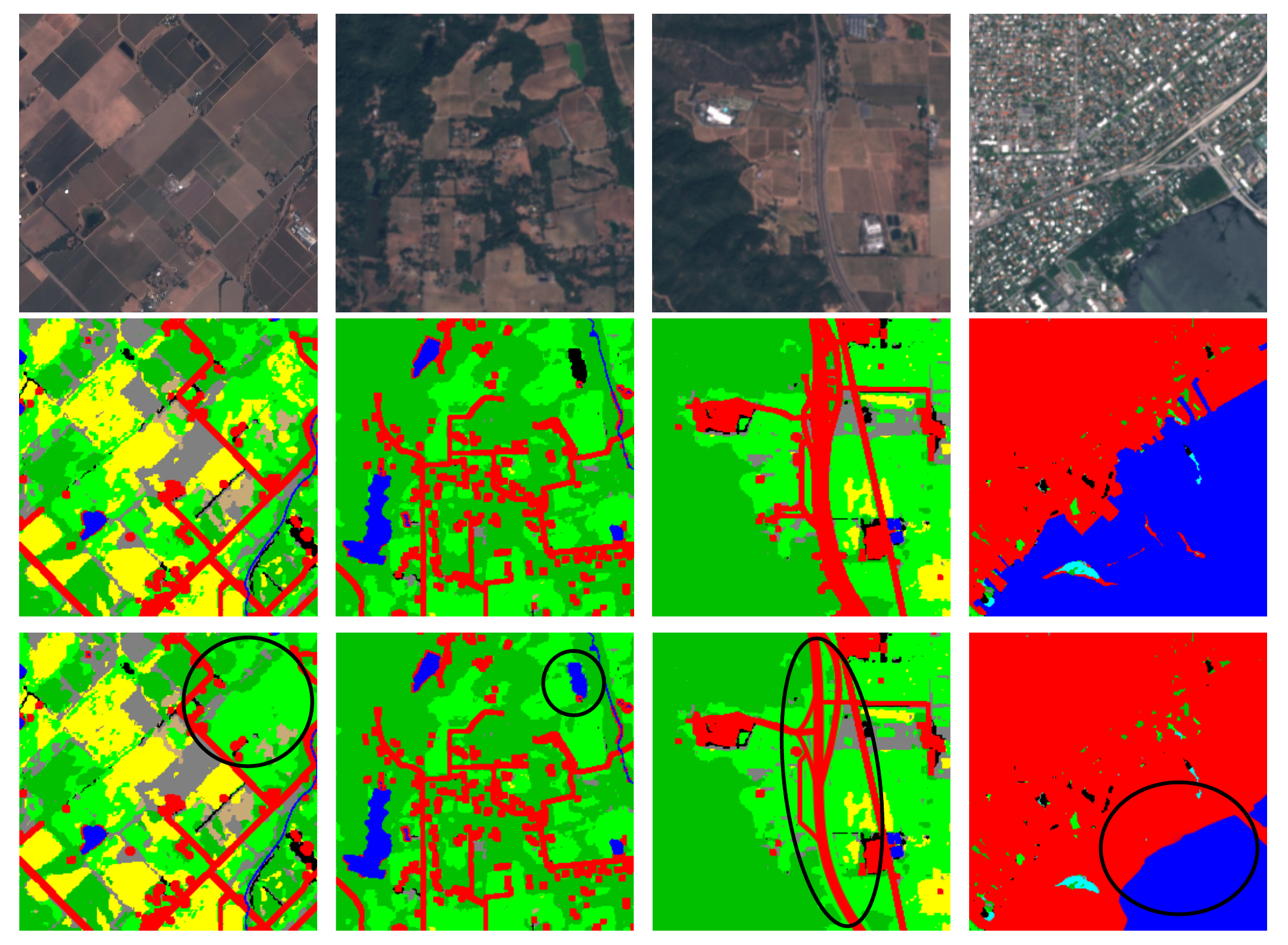}
    \end{center}
    \vspace{-20pt}
    \caption{Examples of human expert verification of mask accuracy. The first row shows original images, the second row shows original masks, and the third row shows masks modified by human annotators.}
     \vspace{-4mm}
    \label{fig:lowquality}
\end{figure*}

\begin{figure*}[h]
    \vspace*{-1.5em} 
    \begin{center}
        \includegraphics[width=\textwidth]{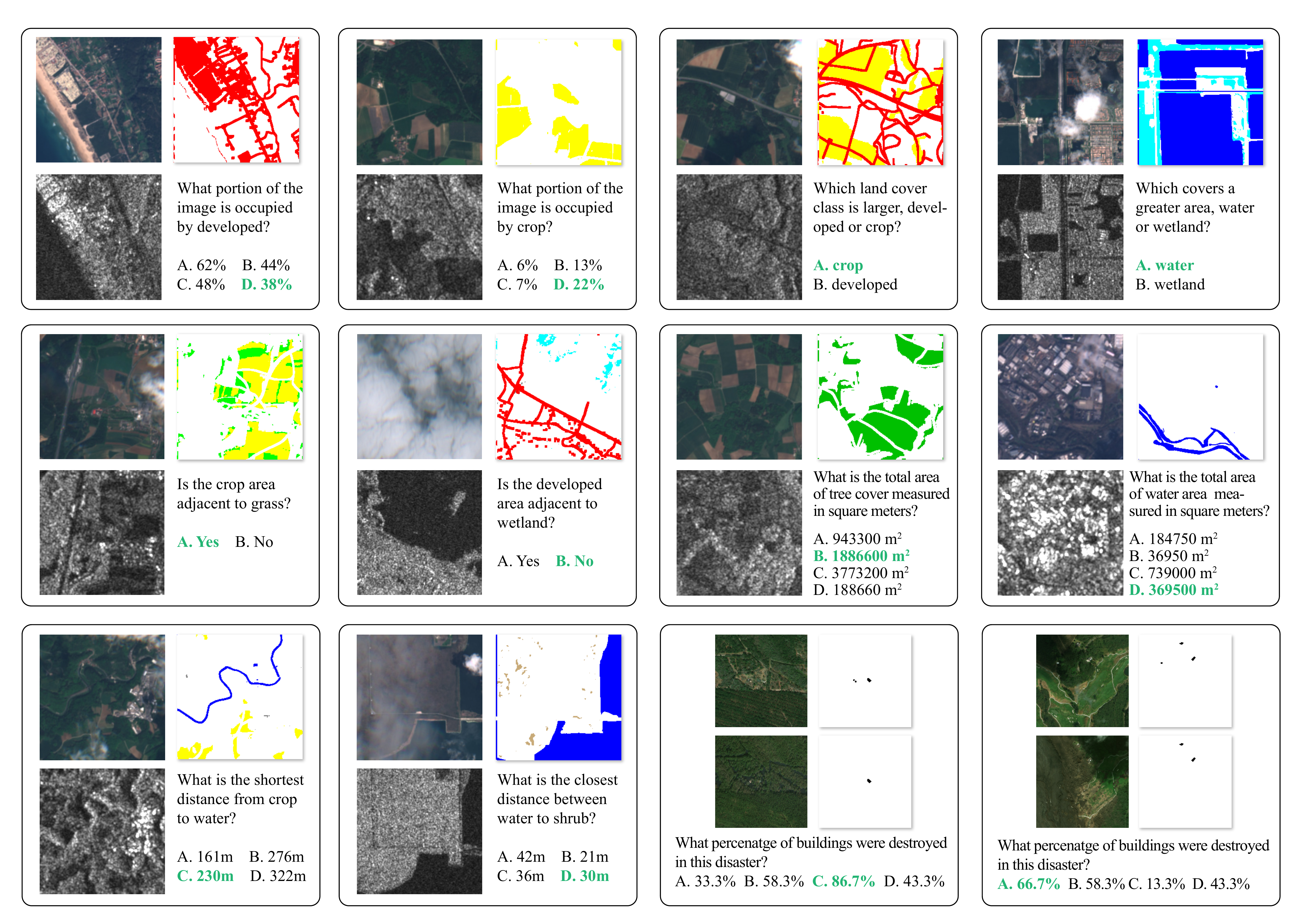}
    \end{center}
    \vspace{-20pt}
    \caption{More examples of TerraScope-Bench, including the questions, answers and the masks involved in the CoT.}
     \vspace{-4mm}
    \label{fig:morebenchmark}
\end{figure*}

\noindent\textbf{Stage 1: Heuristic-Based Answer Generation.}
We leverage existing pixel-level segmentation annotations to generate ground-truth 
answers using deterministic rules. The benchmark includes three data sources: 
ChatEarthNet and BigEarthNet for land cover analysis and xBD for building damage assessment. 
For each image, we process the segmentation mask to extract spatial information 
required for different task types. The specific rules for each task are:

\begin{itemize}[leftmargin=*]
    \item \textbf{Absolute Area Calculation}: For a given land cover class 
    $c$, we count all pixels with label $c$ in the segmentation mask. The 
    area is computed as $A_c = N_c \times r^2$, where $N_c$ is the pixel 
    count and $r$ is the spatial resolution (10m for Sentinel-2). Questions 
    specify a single target class (e.g., ``What is the area of forest?"), 
    and the ground-truth answer is the computed area in square meters or 
    hectares. We only include classes with $A_c > 0$ to avoid trivial questions.
    
    \item \textbf{Coverage Percentage}: For a target land cover class $c$, 
    we compute the percentage as $P_c = \frac{N_c}{N_{\text{total}}} \times 100\%$, 
    where $N_c$ is the pixel count of class $c$ and $N_{\text{total}}$ is 
    the total number of valid pixels in the image (excluding background/void). 
    Questions ask for the coverage of a specific class (e.g., ``What percentage 
    of the image is cropland?"). We require $P_c \geq 5\%$ to ensure the 
    class is visually significant and avoid questions about negligible regions.
    
    \item \textbf{Comparative Area Ranking}: Given a set of land cover 
    classes present in the image, we rank them by area in descending order: 
    $c_1, c_2, \ldots, c_n$ where $A_{c_1} \geq A_{c_2} \geq \cdots \geq A_{c_n}$. 
    Questions ask for the largest class (e.g., ``Which land cover type has 
    the largest area?") or relative ranking (e.g., ``Is forest larger than 
    grassland?"). We only include classes with $P_c \geq 5\%$ in the ranking 
    to ensure clear visual distinction. For binary questions, we require 
    $|A_{c_i} - A_{c_j}| > 0.1 \times \max(A_{c_i}, A_{c_j})$ to avoid 
    ambiguous comparisons between similar-sized regions.
    
    \item \textbf{Distance Measurement}: To measure the minimum distance 
    between two land cover classes $c_i$ and $c_j$, we: (1) generate binary 
    masks $M_i$ and $M_j$ for each class; (2) apply Euclidean distance 
    transform (\texttt{distance\_transform\_edt}) to $M_i$ to compute the 
    distance from each pixel to the nearest $c_i$ pixel; (3) extract the 
    minimum value within $M_j$, which gives the minimum distance $d(c_i, c_j)$ 
    in pixels; (4) convert to meters using spatial resolution ($d_{\text{meters}} = d \times r$). 
    Questions specify two distinct classes (e.g., ``What is the distance between 
    forest and water?"). We require both classes to form spatially connected 
    components (removing isolated pixels via morphological opening) and 
    enforce $d(c_i, c_j) > 10$ pixels to avoid trivial adjacent cases. 
    For classes with multiple disconnected regions, we report the minimum 
    distance across all region pairs.
    
    \item \textbf{Boundary Relationship Detection}: To determine if two 
    land cover classes $c_i$ and $c_j$ are adjacent, we: (1) generate binary 
    masks $M_i$ and $M_j$; (2) apply morphological dilation (\texttt{binary\_dilation}) 
    with a 3×3 structuring element to $M_i$, creating $M_i^{\text{dilated}}$; 
    (3) check if $M_i^{\text{dilated}} \cap M_j \neq \emptyset$. If the 
    intersection is non-empty, the classes are considered adjacent (sharing 
    a boundary). Questions ask binary yes/no queries (e.g., ``Does forest 
    border water?"). We filter out class pairs where either region is too 
    small ($P_c < 3\%$) or fragmented (more than 5 disconnected components) 
    to ensure clear, unambiguous boundaries. For multi-component classes, 
    adjacency is determined if any component pair satisfies the criterion.
    
    \item \textbf{Building Change Estimation}: Using the xBD dataset, we 
    compare pre-disaster and post-disaster satellite imagery to identify 
    destroyed buildings. The annotation process: (1) parse building footprint 
    polygons from JSON files in WKT format using Shapely (\texttt{wkt.loads}); 
    (2) filter polygons based on damage classification labels (only retain 
    buildings labeled as ``destroyed"); (3) rasterize polygon geometries to 
    binary masks using OpenCV (\texttt{cv2.fillPoly}) at the image resolution; 
    (4) count destroyed buildings $N_{\text{destroyed}}$ and total buildings 
    $N_{\text{total}}$ to compute damage rate $R = \frac{N_{\text{destroyed}}}{N_{\text{total}}} \times 100\%$. 
    Questions ask about building counts (e.g., ``How many buildings were destroyed?") 
    or damage percentages (e.g., ``What percentage of buildings were destroyed?"). 
    We only include samples with $N_{\text{total}} \geq 10$ buildings and 
    $N_{\text{destroyed}} \geq 3$ to ensure statistically meaningful damage 
    assessment. Polygon parsing handles potential coordinate precision issues 
    and self-intersecting geometries using Shapely's built-in validation.
\end{itemize}

The implementation uses Python libraries including NumPy for array operations, 
SciPy for distance transforms (\texttt{distance\_transform\_edt}, 
\texttt{binary\_dilation}), Shapely for geometry processing (\texttt{wkt.loads}, 
\texttt{Polygon}), and OpenCV for mask rendering.

\noindent\textbf{Stage 2: GPT-4o-Based Question Refinement.}
To ensure linguistic diversity and difficulty, we use GPT-4o to: (1) rephrase 
template questions into natural language variations, and (2) generate plausible 
distractors for multiple-choice format. For comparative area ranking and 
boundary relationship detection, we generate 2 options (binary choice). For 
other tasks (absolute area, coverage percentage, distance measurement, building 
change estimation), we generate 4 options. The rephrasing prompt is designed to 
maintain semantic equivalence while varying question structure and wording.

\noindent\textbf{GPT-4o Rephrasing Prompt:}
\begin{tcolorbox}[colback=gray!5,colframe=gray!75,title=Question Rephrasing Prompt]
\small
\texttt{Given the following question template and answer:\\
Question: \{original\_question\}\\
Answer: \{ground\_truth\_answer\}\\
\\
Task: Rephrase the question to make it more natural and diverse while preserving 
the original meaning. Generate \{num\_options\} plausible but incorrect answer 
choices (distractors) that are numerically/semantically close to the ground 
truth but clearly distinguishable. Ensure distractors are realistic and 
challenging.\\
\\
Output format:\\
\{\\
\quad "question": "rephrased question",\\
\quad "options": ["option A", "option B", "option C", "option D"],\\
\quad "answer": "correct option letter"\\
\}}
\end{tcolorbox}

\noindent\textbf{Stage 3: Expert Validation.}
We recruit 4 domain experts in geoscience and disaster assessment to ensure 
annotation quality. Each expert is assigned to validate one or two specific 
task types. The validation process includes:
\begin{enumerate}[leftmargin=*]
    \item \textbf{Mask accuracy check}: Verify that segmentation masks 
    correctly represent land cover boundaries or building footprints, as shown in Fig. \ref{fig:lowquality}.
    \item \textbf{Answer correctness}: Validate that ground-truth answers 
    match the mask through manual calculation.
    \item \textbf{Distractor quality}: Ensure distractors are plausible but 
    clearly incorrect.
    \item \textbf{Question clarity}: Check that questions are unambiguous 
    and answerable from the image.
\end{enumerate}

After initial annotation, experts cross-validate each other's work and score 
sample quality on a 3-point scale (low/medium/high). Only samples with 
consensus (all experts agree on high quality) are retained. Samples with 
erroneous masks, ambiguous questions, or invalid distractors are filtered out. 
The final benchmark contains 3,837 expert-verified samples across six task 
types. Sample visualizations are shown in Fig.~\ref{fig:morebenchmark}.

\section{Details of Training Data}
\label{appendix:traindata}

\subsection{Pretraining Data}
For Stage 1 grounded pretraining, we synthesize 2M referring expression 
segmentation (RES) samples from two sources: 1.5M from BigEarthNet
and 0.5M from ChatEarthNet. Both datasets provide semantic 
segmentation annotations with pixel-level class labels. To convert them into 
RES format, we randomly select one land cover category from each image and 
construct the instruction as ``Please segment the [class name]'', where [class name] 
is replaced with the specific land cover type (e.g., ``forest'', ``cropland'', 
``water''). The corresponding ground-truth masks are extracted from the original 
semantic labels and encoded in Run-Length Encoding (RLE) format for efficient 
storage. This synthetic RES data enables the mask decoder to learn foundational 
pixel-level grounding capabilities before instruction tuning.

\subsection{Terra-CoT Dataset Construction}
\label{appendix:terra_cot}

\noindent\textbf{Cap-CoT Curation.} 
We construct the Cap-CoT (Caption with Chain-of-Thought) dataset from four 
sources: ChatEarthNet, BigEarthNet, xBD, and TEOChat (region-based change 
question answering). We employ an RoI-based summarization strategy where 
class information or original metadata, along with mask-overlaid images, are 
fed into Qwen3-VL-235B to generate captions with reasoning chains. The 
generation prompt instructs the model to produce chain-of-thought reasoning 
that explicitly refers to the provided segmentation semantic labels. This 
ensures that generated captions are grounded in precise spatial information 
rather than vague descriptions.

\begin{tcolorbox}[colback=gray!5,colframe=gray!75,title=Caption Generation Prompt for Cap-CoT,breakable]
\small
\textbf{System:} You are an expert in remote sensing image analysis. Your task 
is to generate a detailed caption with step-by-step reasoning for the given 
satellite image.

\vspace{0.3em}
\textbf{Input:}
\begin{itemize}[nosep,leftmargin=1em]
    \item Satellite image with mask overlay
    \item Segmentation labels: \{label\_1, label\_2, ..., label\_n\}
    \item Metadata: [resolution, sensor type, location]
\end{itemize}

\vspace{0.3em}
\textbf{Instructions:}
\begin{enumerate}[nosep,leftmargin=1.5em]
    \item Analyze the spatial distribution of each land cover type shown in 
    the segmentation masks
    \item Generate a chain-of-thought reasoning process that:
    \begin{itemize}[nosep]
        \item Explicitly mentions each segmented region
        \item Describes spatial relationships between different land cover types
        \item Estimates approximate coverage or area for major land cover classes
        \item Notes any significant patterns or features
    \end{itemize}
    \item Provide a final comprehensive caption summarizing the image
\end{enumerate}

\vspace{0.3em}
\textbf{Output Format:}\\
\texttt{<think>}\\
\texttt{First, I observe [description of dominant land cover]. The segmentation 
shows [specific area/pattern]. [SEG for region 1] covers approximately [percentage/area]. 
Next, I notice [another land cover type]. [SEG for region 2] appears in 
[location/pattern]. The spatial relationship between these regions shows 
[description]. Additionally, [other observations]...}\\
\texttt{</think>}\\
\texttt{<caption>}\\
\texttt{This satellite image shows [comprehensive summary including all major 
land cover types, their spatial distribution, and key characteristics].}\\
\texttt{</caption>}
\end{tcolorbox}

\noindent\textbf{VQA-CoT Curation.} 
Based on the 250K Cap-CoT dataset, we first train TerraScope-Cap, a 
caption-specialized variant of TerraScope. We then use TerraScope-Cap to 
annotate images from ChatEarthNet, BigEarthNet, RSVQA-LR, and xBD training 
sets, generating captions and predicted masks. For ground-truth mask refinement, 
we compute the intersection between predicted masks and available ground-truth 
annotations when available, ensuring higher quality.

Using these captions as context, we synthesize L1-level VQA samples covering 
six task types. We design predefined templates for each task type to ensure 
consistency and coverage:

\begin{tcolorbox}[colback=blue!5,colframe=blue!75,title=L1-Level VQA Templates,breakable]
\small
\textbf{Task 1: Object Existence}\\
Template: ``Is there any [class] in the image?''\\
Example: ``Is there any forest in the image?''\\
Answer: ``Yes'' or ``No''

\vspace{0.5em}
\textbf{Task 2: Object Counting}\\
Template: ``How many [object] are there in the image?''\\
Example: ``How many buildings are there in the image?''\\
Answer: ``[number] [object]'' (e.g., ``15 buildings'')

\vspace{0.5em}
\textbf{Task 3: Localization}\\
Template: ``Where is the [class] located in the image?''\\
Example: ``Where is the water body located in the image?''\\
Answer: ``[cardinal direction/relative position]'' (e.g., ``in the northeastern 
part'', ``along the southern edge'')

\vspace{0.5em}
\textbf{Task 4: Area Quantification}\\
Template 1: ``What is the area of [class]?''\\
Template 2: ``What percentage of the image is covered by [class]?''\\
Example 1: ``What is the area of cropland?''\\
Example 2: ``What percentage of the image is covered by forest?''\\
Answer 1: ``[number] square meters'' or ``[number] hectares''\\
Answer 2: ``[percentage]\%''

\vspace{0.5em}
\textbf{Task 5: Boundary Detection}\\
Template: ``Does [class1] border [class2]?''\\
Example: ``Does forest border water?''\\
Answer: ``Yes'' or ``No''

\vspace{0.5em}
\textbf{Task 6: Distance Measurement}\\
Template: ``What is the distance between [class1] and [class2]?''\\
Example: ``What is the distance between cropland and water?''\\
Answer: ``[number] meters''

\vspace{0.5em}
\textbf{Generation Strategy:}
\begin{itemize}[nosep,leftmargin=1em]
    \item For each image, randomly select 2-4 task types
    \item Ensure at least one task per image requires pixel-level reasoning
    \item Classes are sampled from available segmentation labels
    \item Answers are computed deterministically from ground-truth or refined masks
\end{itemize}
\end{tcolorbox}

Building upon L1-level VQA, we use GPT-4o to synthesize more complex reasoning 
problems that require multi-step spatial analysis. The synthesis prompt 
encourages GPT-4o to create questions involving comparative reasoning, spatial 
relationships, and compositional understanding.

\begin{figure*}[t]
    \vspace*{-2mm} 
    \begin{center}
        \includegraphics[width=\textwidth]{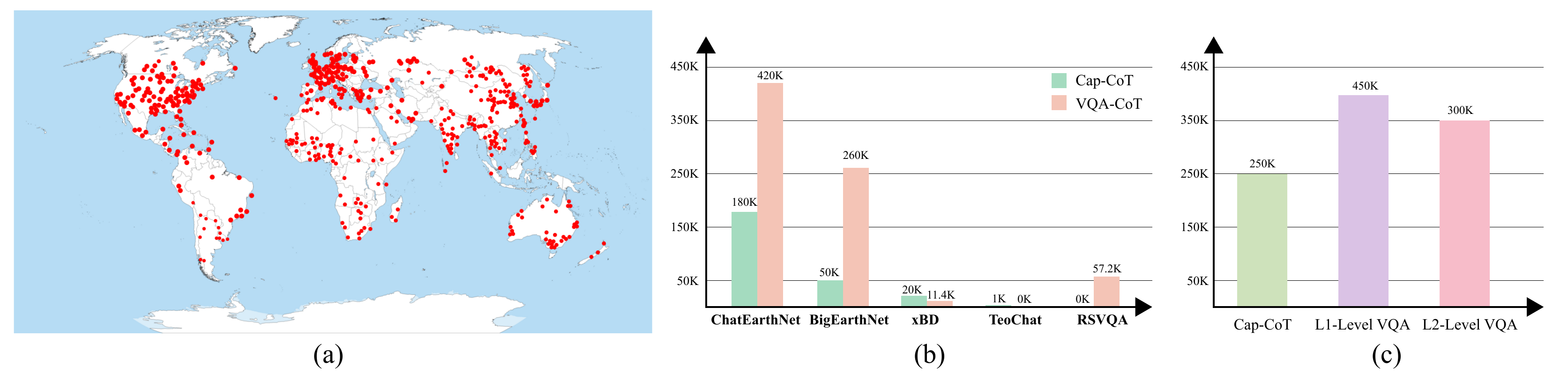}
    \end{center}
    \vspace{-20pt}
    \caption{Data distributions of Terra-CoT.}
     \vspace{-4mm}
    \label{fig:train_sta}
\end{figure*}

Fig.~\ref{fig:train_sta} visualizes the composition and distribution of the 
Terra-CoT dataset from three perspectives. First, we show the geographic 
distribution of source images, 
demonstrating global coverage across diverse geographical regions and climatic 
zones. Second, we present the data source breakdown for Cap-CoT and VQA-CoT 
subsets, illustrating how different source datasets contribute to caption 
generation and question-answering components. Third, we provide sample quantity 
statistics across the three dataset tiers: Cap-CoT (caption with chain-of-thought), 
L1-level VQA (simple spatial queries), and L2-level VQA (complex multi-step 
reasoning).

\begin{tcolorbox}[colback=green!5,colframe=green!75,title=L2-Level VQA Synthesis Prompt,breakable]
\small
\textbf{System:} You are an expert in designing complex spatial reasoning 
questions for satellite imagery analysis.

\vspace{0.3em}
\textbf{Input:}
\begin{itemize}[nosep,leftmargin=1em]
    \item Image caption with spatial information
    \item L1-level QA pairs (simple questions and answers)
    \item Available land cover classes: \{class\_1, class\_2, ..., class\_n\}
\end{itemize}

\vspace{0.3em}
\textbf{Task:} Generate 2-3 complex reasoning questions from two categories:

\vspace{0.3em}
\textbf{Category 1: Spatial Reasoning Questions}

These questions focus on geometric and spatial properties requiring pixel-level 
analysis, such as area comparison, distance measurement, boundary relationships, 
coverage quantification, and spatial distribution patterns.

\vspace{0.3em}
\textbf{Category 2: Semantic Reasoning Questions}

These questions focus on understanding land cover semantics, ecological patterns, 
temporal changes, functional relationships, and overall landscape composition.

\vspace{0.3em}
\textbf{Requirements:}
\begin{enumerate}[nosep,leftmargin=1.5em]
    \item Generate at least one question from each category
    \item Questions must require multi-step reasoning
    \item Answers should be deterministic and verifiable
    \item Spatial reasoning questions must involve precise geometric analysis
    \item Semantic reasoning questions must demonstrate understanding of land 
    cover semantics
\end{enumerate}

\vspace{0.3em}
\textbf{Output Format:}\\
For each question, provide:
\begin{itemize}[nosep,leftmargin=1em]
    \item Question text
    \item Category: [Spatial Reasoning] or [Semantic Reasoning]
    \item Ground-truth answer
    \item Reasoning steps required (brief description)
    \item Classes involved
\end{itemize}
\end{tcolorbox}

\begin{table}[h]
\centering
    \centering
   
    \vspace{-0.1in}
    \renewcommand{\arraystretch}{1.15}

    \begin{tabular}{>{\kern-0.5\tabcolsep}l|c<{\kern-0.5\tabcolsep}}
        \toprule
        \textbf{Hyperparameter}  & \textbf{Value}   \\
        \midrule
         Overall batch size  &32  \\
        Learning rate  & 4e-5 \\
        LR Scheduler  & Cosine decay \\
        DeepSpeed ZeRO Stage &ZeRO-2 \\
        Optimizer & Adam \\
        Warmup ratio & 0.3 \\
        Epoch & 1 \\
        Weight decay & 0 \\
        Precision & bf16 \\ 
        \bottomrule
    \end{tabular}
     \caption{Hyperparameters of TerraScope.}
    \label{tab:hyper}
\end{table}

\section{Experimental Settings}
\label{appendix:settings}

\subsection{Training Details of TerraScope}
\label{appendix:training_details}

We provide complete training hyperparameters for TerraScope in 
Tab.~\ref{tab:hyper}. The model is fine-tuned for one epoch on Terra-CoT 
with a total batch size of 32, using the Adam optimizer with cosine learning 
rate decay. We employ DeepSpeed ZeRO-2 for memory-efficient training and 
use bf16 mixed precision to accelerate computation.


\noindent\textbf{System Prompt for Reasoning Mode.}
To facilitate pixel-grounded reasoning, we modify the system prompt of 
InternVL-3 to encourage structured thinking with segmentation. The modified 
prompt instructs the model to: (1) generate reasoning steps within 
\texttt{<think>} tags, where it can insert \texttt{[SEG]} tokens to trigger 
mask generation; (2) provide the final answer within \texttt{<answer>} tags. 
The complete system prompt is:

\begin{tcolorbox}[colback=gray!5,colframe=gray!75,title=System Prompt for TerraScope]
\small
\texttt{A conversation between User and Assistant. The user asks a question, 
and the Assistant solves it. The Assistant first thinks about the reasoning 
process in their mind, generating segmentation masks when needed using [SEG] 
tokens, and then provides the user a concise final answer in a short word or 
phrase. The reasoning process and answer are enclosed within <think> </think> 
and <answer> </answer> tags, respectively, i.e., <think> reasoning process 
with [SEG] for segmentation </think><answer> answer here </answer>.}
\end{tcolorbox}

\noindent\textbf{Example Prompt with Pixel-Grounded Reasoning:}

\begin{tcolorbox}[colback=blue!5,colframe=blue!75,title=Example Output,breakable]
\small
\textbf{User:} What percentage of the image is covered by forest?

\vspace{0.5em}
\textbf{Assistant:} \\
\texttt{<think>}\\
To answer this question, I need to identify and measure the forest area in 
the image. Let me first segment the forest regions. [SEG]\\
\\
\textit{[Mask generated for forest regions]}\\
\\
Now I can see the forest areas clearly marked. Let me calculate the coverage:\\
- The segmented forest region covers approximately 3,450 pixels\\
- The total image area is about 10,000 pixels\\
- Coverage percentage = (3,450 / 10,000) × 100\% = 34.5\%\\
\\
Therefore, forest covers approximately 35\% of the image.\\
\texttt{</think>}\\
\\
\texttt{<answer>}35\%\texttt{</answer>}
\end{tcolorbox}

The full prompt template including additional few-shot examples is provided 
in our code repository.

\subsection{Evaluation Details on TerraScope-Bench}
\label{appendix:eval_details}

\noindent\textbf{Baseline Models.}
We evaluate TerraScope-Bench on three categories of vision-language models:

\begin{itemize}[leftmargin=*]
    \item \textbf{General-purpose VLMs}: We include proprietary models 
    (GPT-4o) and open-source models (LLaVA-OV), 
    Qwen2.5-VL, InternVL-3. For GPT-4o, 
    we utilize its multi-image API to handle multi-temporal inputs for the 
    Building Change Estimation task.
    
    \item \textbf{Reasoning-enhanced VLMs}: We evaluate two models with 
    explicit reasoning capabilities: GLM-4.1V-Think and 
    Qwen3-VL-Think, which generate intermediate reasoning 
    steps before final answers.
    
    \item \textbf{EO-specific VLMs}: We include domain-adapted models 
    fine-tuned on earth observation data: GeoChat, 
    TEOChat, LHRS-Bot, 
    EarthDial, and EarthMind.
\end{itemize}

\noindent\textbf{Evaluation Protocol.}
All tasks in TerraScope-Bench are formatted as multiple-choice questions 
with 2 or 4 options (A, B, C, D). We use a unified prompt template across 
all evaluated models, requesting them to select the correct option. To ensure 
reliable option extraction, we incorporate \textbf{option prediction guidance} 
in the prompt: \textit{"Please respond with only the option letter (A, B, C, 
or D) corresponding to your answer."} Since some models have limited 
instruction-following ability and may generate verbose explanations instead 
of direct option letters, we implement post-processing using regex patterns 
(e.g., \texttt{r'\textbackslash b[A-D]\textbackslash b'}) to extract the 
predicted option from model outputs. If multiple option letters appear, we 
select the first occurrence; if no valid option is found, the prediction is 
marked as incorrect.

\noindent\textbf{Multi-temporal Handling.}
For the Building Change Estimation task, which requires comparing pre-disaster 
and post-disaster imagery:
\begin{itemize}[leftmargin=*,nosep]
    \item Proprietary models (GPT-4o): Use multi-image input API
    \item Open-source models: Concatenate images horizontally or process 
    as separate frames
    \item Models without multi-image support: Provide both images sequentially 
    in the conversation
\end{itemize}

\noindent\textbf{Evaluation Metrics.}
We compute accuracy by exact matching between predicted option letters and 
ground-truth answers. For each task type, we report:
\begin{itemize}[leftmargin=*,nosep]
    \item Per-task accuracy: Percentage of correct predictions for each task
    \item Overall accuracy: Macro-average across all six tasks
\end{itemize}

\noindent\textbf{Implementation Details.}
\begin{itemize}[leftmargin=*,nosep]
    \item For open-source models, we use their official repositories and 
    recommended inference settings
    \item For proprietary APIs (GPT-4o), we set temperature=0 for deterministic 
    outputs
    \item All evaluations use greedy decoding (top-p=1.0, temperature=0) 
\end{itemize}

To ensure fair comparison, we fine-tune baseline models on our Terra-CoT 
dataset with appropriate adaptations:

\begin{itemize}[leftmargin=*]
    \item \textbf{InternVL-3}: We remove all special tokens (\texttt{<think>}, 
    \texttt{</think>}, \texttt{[SEG]}) from the training data and perform 
    standard supervised fine-tuning using the official training scripts. 
    The model is trained to directly predict answers without explicit 
    reasoning traces or segmentation masks.
    
    \item \textbf{GLM-4.1V-Think}: We preserve the thinking mode structure 
    (\texttt{<think>}, \texttt{</think>}) but remove the \texttt{[SEG]} 
    token, as this model does not support pixel-level grounding. We use 
    the official training pipeline combining SFT (Supervised Fine-Tuning) 
    and RLVR (Reinforcement Learning with Verifiable Rewards) as described 
    in~\cite{hong2025glm}.
\end{itemize}

This design allows us to assess whether baseline models can benefit from 
our training data while maintaining their original architectures. The complete evaluation code, prompts, and output parsing scripts are available 
in our repository.

\section{Efficiency Analysis}
\label{appendix:efficiency}

We analyze TerraScope's computational efficiency from multiple perspectives, 
including inference time, memory consumption, parameter count, and the impact 
of pixel-grounded reasoning on computational cost.

\subsection{Model Complexity}
\label{appendix:model_complexity}

Tab.~\ref{tab:model_complexity} compares TerraScope with mainstream baseline 
models in terms of model size.

\begin{table}[h]
\centering
 \resizebox{\linewidth}{!}{
\begin{tabular}{lcc}
\toprule
Model & Total Params & Additional Modules \\
\midrule
GPT-4o & - & - \\
Qwen2.5-VL-7B & 7.6B & - \\
InternVL-3-8B & 8.1B & - \\
GLM-4.1V-9B & 9.4B & - \\
LLaVA-OV-7B & 7.2B & - \\
\midrule
\textbf{TerraScope-8B} & \textbf{8.3B} & \textbf{SAM-2 (0.228B)} \\
\quad -- Base InternVL-3 & 8.1B & - \\
\quad -- SAM-2 image encoder & - & 0.224B \\
\quad -- SAM-2 mask decoder & - & 0.004B \\
\bottomrule
\end{tabular}
}
\caption{Model complexity comparison. TerraScope adds a lightweight pixel-level grounding module on top of InternVL-3.}
\label{tab:model_complexity}
\end{table}

TerraScope integrates the SAM-2 image encoder (224.4M parameters) and mask
decoder (3.9M parameters) to enable pixel-level grounding. These two modules
together introduce only about 0.228B additional parameters, increasing the
overall model size from 8.1B (base InternVL-3) to 8.3B. This corresponds to
a parameter overhead of merely $\sim$2.8\%.

Crucially, the added segmentation components are extremely lightweight compared
to the backbone large multimodal model: the extra 0.228B parameters account for
only a small fraction of the total parameter budget, while the vast majority of
parameters still reside in the LLM. In other words, TerraScope incurs only a
minimal parameter increase yet gains the substantial benefit of being able to
produce verifiable, pixel-level segmentation masks at each reasoning step.

\subsection{Inference Time Analysis}
\label{appendix:inference_time}

We measure inference time on a single NVIDIA A100 80GB GPU with batch size 1. 
Tab.~\ref{tab:inference_time} reports the average time per sample on 
TerraScope-Bench.

\begin{table}[h]
\centering
\begin{tabular}{lc}
\toprule
Model & Avg. Time (s) \\
\midrule
InternVL-3-8B & 0.85 \\
Qwen2.5-VL-7B & 0.92 \\
\midrule
\textbf{TerraScope-8B} & \textbf{2.48} \\
GLM-4.1V-9B & 2.60 \\
\bottomrule
\end{tabular}
\caption{Average inference time per sample (seconds).}
\label{tab:inference_time}
\end{table}

TerraScope achieves faster inference than GLM-4.1V-9B (2.4s vs 2.6s) despite 
generating additional segmentation masks. We identify two key efficiency 
advantages: First, TerraScope performs deterministic reasoning with structured 
output (\texttt{<think>} and \texttt{<answer>} tags), while GLM-4.1V tends 
to generate overly verbose reasoning traces with significantly more tokens. 
Second, our interleaved mask injection is highly efficient—masked visual 
features are directly inserted into the KV cache without re-encoding through 
the vision encoder, avoiding redundant visual processing. InternVL-3 remains 
the fastest (0.85s) as it generates answers directly without reasoning, but 
lacks both reasoning transparency and pixel-level grounding capabilities that 
TerraScope provides.

\subsection{Memory Consumption}
\label{appendix:memory}

We profile GPU memory usage during inference on a single NVIDIA A100 80GB GPU. 
Tab.~\ref{tab:memory} shows peak memory consumption with different numbers 
of generated masks.

\begin{table}[h]
\centering
\begin{tabular}{lccc}
\toprule
Model & 1 Mask & 2 Masks & 3+ Masks \\
\midrule
InternVL-3-8B & 18.2 & 18.3 & 18.2 \\
Qwen2.5-VL-7B & 16.8 & 17.0 & 17.0 \\
\midrule
\textbf{TerraScope-8B} & \textbf{22.4} & \textbf{23.1} & \textbf{24.2} \\
\bottomrule
\end{tabular}
\caption{GPU memory consumption (GB) on NVIDIA A100.}
\label{tab:memory}
\end{table}

TerraScope requires approximately 22\% more memory than InternVL-3 (22.4GB vs 
18.2GB for single-mask cases), primarily due to the SAM-2 decoder weights 
(3.9GB). Memory consumption increases with the number of generated masks, 
as each mask adds approximately 0.7GB for storing mask features and intermediate 
activations. In contrast, baseline models (InternVL-3, Qwen2.5-VL) maintain 
constant memory usage regardless of output complexity, as they do not generate 
pixel-level grounding. The memory overhead is acceptable given TerraScope's 
additional capability of producing verifiable segmentation masks.

\begin{table*}[ht]                 
  \centering
  \small
  \setlength{\tabcolsep}{5pt}
  \begin{tabular}{@{}lccccccccccc@{}}
    \toprule
    Model  & Size & APR & COA & DLC & FOD&
           MOP & NUM& SRI & USR & Overall \\
    \midrule
    EarthDial  & 4B & 23.49 & 10.34 & 75.27 & \textbf{99.00} & 61.16 & 43.62 & 51.24 & 15.52 & 48.29 \\ 
    RS-LLaVA   & 7B & 68.57 & 80.88 & 71.24 & 87.00 & 63.09 & 49.85 & 26.17 & 10.34 & 57.24 \\
    MiMo       & 7B & 40.00 & 45.77 & 92.47 & 93.33 & \textbf{84.30} & \textbf{61.42} & \textbf{94.21}& \textbf{88.97} & 75.55 \\ 
    GLM-4.1V      & 9B & 45.71 & 36.36 & 72.85 & 62.67 & 67.49 & 58.63 & 69.97 & 88.28 & 62.87 \\ 
    Qwen2.5-V       & 7B & 29.84 & 89.66 & \textbf{94.09} & 71.67 & 76.03 & 53.12 & 92.84 & 82.07 & 74.28 \\
    LLaVA-OV      & 8B & 39.37 & 79.00 & 83.06 & 59.00 & 72.45 & 46.59 & 85.12 & 10.34 & 60.96 \\
    \midrule
    TerraScope & 8B & \textbf{69.84} & \textbf{98.12} & 83.06 & 87.67 & 61.98 & 60.82 & 91.12 & 85.52 & \textbf{79.36} \\
    \bottomrule
  \end{tabular}

    \caption{Performance on the VQA task on Landsat30-AU. Bold indicates the best score.}
      \label{tab:compre_landsat}
\end{table*}

\section{Additional Experimental Results}
\label{appendix:additional_results}

Beyond the geospatial reasoning tasks reported in Sec.~\ref{sec:experiments}, 
we evaluate TerraScope on additional benchmarks to demonstrate its generalization 
ability across diverse earth observation tasks.

\subsection{Comprehensive Results on Landsat30-AU}
\label{appendix:landsat}

Tab.~\ref{tab:compre_landsat} presents complete results on all eight task 
types in Landsat30-AU. The benchmark includes Agro-Phenology 
Reasoning (APR) for agricultural growth stages, Cloud-Occlusion Assessment (COA) 
for detecting cloud coverage, Dominant Land-Cover (DLC) for identifying main 
land types, Fine-Object Detectability (FOD) for detecting small objects, 
Macro-Object Presence (MOP) for large-scale objects, Object Counting (NUM), 
Spatial Relationship (SRI) for spatial layout reasoning, and Urban Scale 
Recognition (USR) for classifying settlement scale. TerraScope achieves competitive performance across all task types, with 
particularly strong results on fine-grained visual tasks requiring precise 
spatial understanding, such as Cloud-Occlusion Assessment (COA) and Fine-Object 
Detectability (FOD). This demonstrates that pixel-grounded reasoning capabilities 
transfer effectively to general earth observation understanding tasks.

\subsection{Results on RSVQA and Scene Classification}
\label{appendix:rsvqa_scene}

Tab.~\ref{tab:rsvqa_scene} reports performance on RSVQA-LR~\cite{lobry2020rsvqa} and 
BigEarthNet scene classification~\cite{sumbul2019bigearthnet}. On RSVQA-LR, TerraScope 
performs slightly below  EarthDial. We attribute this to the 
difference in training data scale—LHRS-Bot and EarthDial were trained on 
significantly larger VQA datasets, which benefits 
general question-answering tasks. On BigEarthNet scene classification, TerraScope 
achieves competitive accuracy comparable to EarthDial, demonstrating effective 
transfer learning despite being primarily designed for pixel-grounded reasoning.

\begin{table}[h]
\centering
\begin{tabular}{lcc}
\toprule
Model & RSVQA-LR & BigEarthNet \\
\midrule
GeoChat & 90.7 & 20.4 \\
LHRS-Bot &89.2 & - \\
EarthDial & \textbf{92.7} & 68.8 \\
\midrule
\textbf{TerraScope} & 91.4 & \textbf{69.2} \\
\bottomrule
\end{tabular}
\caption{Results on RSVQA and scene classification.}
\label{tab:rsvqa_scene}
\end{table}

\subsection{Complete Results on DisasterM3}
\label{appendix:disasterm3}

We report comprehensive results on DisasterM3, which includes 
both optical-optical and optical-SAR multi-modal evaluation. In the main paper 
(Sec.~\ref{sec:experiments}), we reported only optical-optical results as most 
baseline models do not support SAR imagery. Tab.~\ref{tab:disasterm3_full} 
presents results on both modality configurations. TerraScope is the only model 
capable of handling optical-SAR multi-modal inputs through adaptive modality 
selection. On optical-optical pairs, TerraScope achieves competitive performance 
with EO-specific baselines. On optical-SAR pairs, TerraScope demonstrates its 
unique capability to leverage complementary information from heterogeneous 
modalities for damage assessment.

\begin{table}[h]
\centering
\begin{tabular}{lcccc}
\toprule
\multirow{2}{*}{Model} & \multicolumn{3}{c}{Optical-SAR} \\
\cmidrule(lr){2-4}
 & BDC & DRE & Avg \\
\midrule
LLaVA-OV & 22.2 & 19.4 & 20.8 \\
TEOChat & 18.4 & 9.4 & 13.9 \\
InternVL3-8B & 20.7 & 18.4 & 19.6 \\
EarthDial & 19.5 & 10.2 & 14.9 \\
\midrule
\textbf{TerraScope} & \textbf{50.4} & \textbf{32.6} & \textbf{41.5} \\
\bottomrule
\end{tabular}
\caption{Optical-SAR results on DisasterM3 benchmark.}
\label{tab:disasterm3_full}
\end{table}

\begin{table}[t]
    \resizebox{0.98\linewidth}{!}{
    \renewcommand{\arraystretch}{1.15}
    \begin{tabular}{c|ccc}
        \toprule
        \textbf{Data}  &\textbf{TerraBench.} & \textbf{Landsat.} & \textbf{Disaster.} \\
        \midrule      
        Cap-CoT & 42.8& 50.1 & 26.9  \\
        Cap-CoT + L1-VQA & 66.7 & 61.0 & 46.2  \\
        \textbf{Cap-CoT + L1-VQA  + L2-VQA} & \textbf{68.9} & \textbf{73.9} & \textbf{46.5} \\
        \bottomrule
    \end{tabular}
    }
    \vspace{-0.1in}
     \caption{Ablations of Terra-CoT.}
     \label{tab:terra_cot}
\end{table}

\section{More Ablation Studies}
\label{appendix:ab}

\noindent\textbf{Effectiveness of Two-Stage Training.} TerraScope employs a 
two-stage training strategy: Stage 1 performs grounded pretraining on 2M 
referring expression segmentation pairs to train the mask decoder, and Stage 2 
applies instruction tuning on Terra-CoT to jointly optimize the projector, 
LLM, and mask decoder. Tab.~\ref{tab:ablation_pretrain} compares models with 
and without Stage 1 pretraining on three benchmarks. The results demonstrate 
that grounded pretraining establishes foundational pixel-level grounding 
capability, which substantially improves performance on pixel-grounded reasoning 
tasks and also benefits general EO understanding and disaster assessment tasks.

\begin{table}[h]
\centering
 \resizebox{0.98\linewidth}{!}{
\begin{tabular}{lccc}
\toprule
Training Strategy & TerraScope-Bench & Landsat30-AU & DisasterM3 \\
\midrule
w/o Grounded Pretrain & 65.4 & 71.8 & 43.0 \\
\textbf{w/ Grounded Pretrain} & \textbf{68.9} & \textbf{73.9} & \textbf{46.5} \\
\bottomrule
\end{tabular}
}
\caption{Ablation study on grounded pretraining.}
\label{tab:ablation_pretrain}
\end{table}

\begin{figure*}[t]
    \vspace*{-2mm} 
    \begin{center}
        \includegraphics[width=\textwidth]{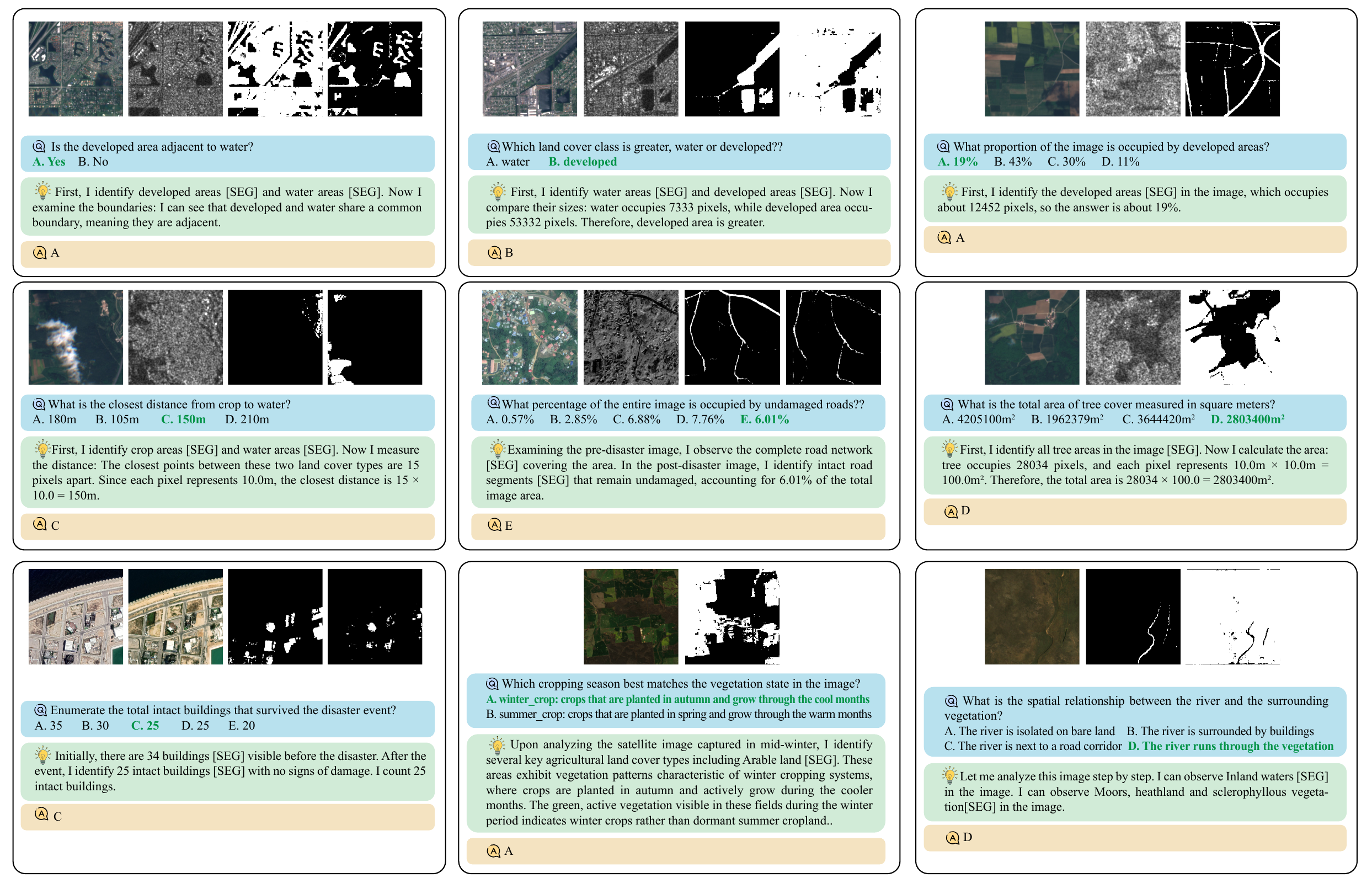}
    \end{center}
    \vspace{-20pt}
    \caption{More qualitative results of TerraScope.}
     \vspace{-4mm}
    \label{fig:morevis}
\end{figure*}

\begin{figure*}[t]
    \vspace*{-2mm} 
    \begin{center}
        \includegraphics[width=\textwidth]{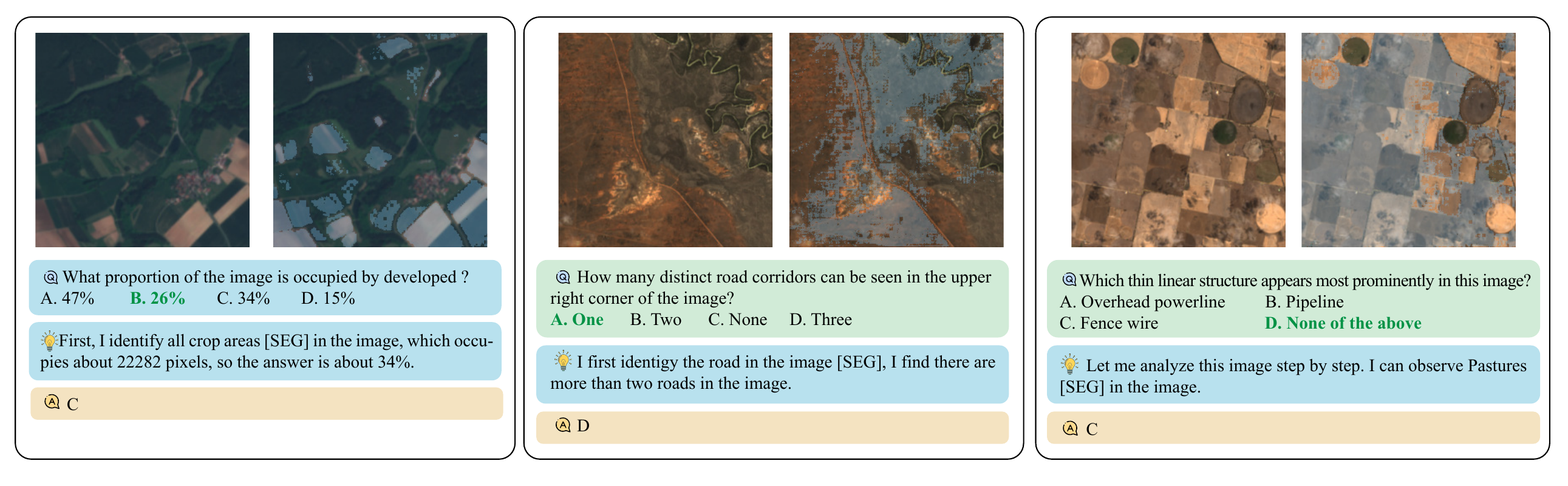}
    \end{center}
    \vspace{-20pt}
    \caption{Failure cases of TerraScope.}
     \vspace{-4mm}
    \label{fig:failure}
\end{figure*}

\noindent\textbf{Effectiveness of Terra-CoT data composition.} Our Terra-CoT dataset 
is synthesized using a hierarchical data synthesis strategy combining three 
data types: L1-level VQA, L2-Level VQA, and captioning. To validate the 
effectiveness of this composition, we train TerraScope with different data 
mixtures in Tab.~\ref{tab:terra_cot}.  First, training with Terra-Cap (captioning only) provides limited instruction-following  capability, as the model struggles with both perception and reasoning tasks. 
Second, adding L1-level VQA establishes foundational pixel-grounded visual 
understanding, significantly improving performance on tasks requiring accurate 
segmentation. However, this perception-focused training still lacks complex 
reasoning capabilities, resulting in poor performance on challenging tasks 
like those in LandSat30-AU that require multi-step spatial reasoning. Third, incorporating L2-Level data enables strong generalization across 
diverse task types. The full Terra-CoT mixture achieves the best overall 
performance, with improvements scaling consistently as we increase the proportion 
of reasoning data.

\noindent\textbf{Ablations about multi-modal reasoning.} We investigate how multi-modal data (optical and SAR) contributes to TerraScope's 
performance. We design ablation experiments by controlling two aspects: 
(1) \textbf{Multi-modal encoding}: whether to concatenate optical and SAR 
features as input to the LLM during initial image encoding; (2) \textbf{Masked 
feature interleaving}: how to inject masked visual features during reasoning 
steps—using optical only, concatenating both modalities, or adaptively selecting 
based on relevance scores (Eq.~4-5).

Tab.~\ref{tab:ablation_multimodal} presents results on TerraScope-Bench, 
evaluated on both segmentation quality (mean IoU) and final answer accuracy.

\begin{table}[h]
\centering
\resizebox{\linewidth}{!}{
\begin{tabular}{lcc|ccc}
\toprule
Multi-modal & Masked Feature & \multicolumn{2}{c}{TerraScope-Bench} & Efficiency \\
Encoding & Interleaving & Mean IoU (\%) & Accuracy (\%) & \\
\midrule
Optical only & Optical only & 53.4 & 65.0 & High \\
Optical only & Concat Opt+SAR & 53.5 & 67.6 & Low \\
Optical only & Adaptive selection & 53.1 & 67.4 & High \\
\midrule
Concat Opt+SAR & Optical only & 56.8 & 69.2 & High \\
Concat Opt+SAR & Concat Opt+SAR & \textbf{57.2} & \textbf{73.0} & Low \\
Concat Opt+SAR & Adaptive selection & \textbf{57.2} & 72.6 & High \\
\bottomrule
\end{tabular}
}
\caption{Ablation study on multi-modal reasoning. "Efficiency" indicates 
inference efficiency: "High" for methods with shorter context length (single 
modality or adaptive selection), "Low" for concatenation methods that double 
the visual token count.}
\label{tab:ablation_multimodal}
\end{table}

Our ablation study reveals two key findings. First, multi-modal encoding is 
essential for both accurate segmentation and reasoning. Concatenating optical 
and SAR features as initial input substantially improves performance compared 
to optical-only encoding, demonstrating that the LLM benefits from complementary 
multi-modal representations from the beginning of reasoning. Second, the 
masked feature injection strategy during reasoning steps also matters. Both 
concatenation and adaptive selection of masked features significantly outperform 
optical-only injection. While concatenation achieves slightly higher answer 
accuracy, adaptive selection demonstrates a favorable trade-off: it maintains 
comparable segmentation quality and nearly equivalent reasoning performance 
while significantly reducing context length by dynamically selecting only the 
most informative modality at each spatial location. This reduction in context 
length translates to substantial savings in memory consumption and inference 
time, making adaptive selection the more practical choice for deployment.

\section{Additional Visualizations and Failure Analysis}
\label{appendix:vis}

\subsection{Qualitative Results}
\label{appendix:vis_success}

Fig.~\ref{fig:morevis} presents additional qualitative results 
demonstrating TerraScope's capabilities across diverse scenarios. The 
visualizations show that TerraScope can perform pixel-grounded reasoning on: 
(1) single-modality optical imagery, generating accurate segmentation masks 
and spatial analysis; (2) multi-modal optical-SAR fusion, adaptively selecting 
the most informative modality for each spatial region; and (3) temporal 
change detection, providing chain-of-thought reasoning traces that explain 
land cover changes with supporting visual evidence. These results validate 
TerraScope's versatility in handling different data modalities and temporal 
information while maintaining pixel-level grounding throughout the reasoning 
process.

\subsection{Failure Cases and Analysis}
\label{appendix:vis_failure}

Fig.~\ref{fig:failure} presents typical failure cases to understand 
TerraScope's limitations. We identify two primary failure modes:

\noindent\textbf{(1) Limited spectral information.} TerraScope currently 
processes only RGB bands as input, discarding additional spectral channels 
available in multispectral sensors like Sentinel-2 (which provides 13 bands 
including near-infrared, red-edge, and shortwave infrared). This limitation 
makes it challenging to distinguish spectrally similar land cover types that 
appear visually identical in RGB but exhibit distinct spectral signatures in 
other bands. For example, certain crop types or vegetation health conditions 
that are easily separable using NDVI or red-edge indices become ambiguous in 
RGB-only input, leading to incorrect segmentation and subsequent reasoning 
errors.

\noindent\textbf{(2) Error propagation from segmentation.} For scenes containing 
small or low-contrast objects (e.g., narrow roads, sparse buildings, thin 
water channels), the mask decoder may produce inaccurate segmentation due to 
insufficient visual salience. These segmentation errors directly propagate to 
the reasoning stage: when spatial claims are grounded in incorrect masks, the 
derived answers become unreliable even if the reasoning logic is sound. This 
highlights the critical dependency of pixel-grounded reasoning on high-quality 
segmentation, particularly for fine-grained objects in complex landscapes.

Future improvements could address these limitations by: (1) extending the 
vision encoder to process full multispectral inputs rather than RGB only, 
enabling better spectral discrimination; (2) incorporating uncertainty estimation 
in the segmentation module to flag low-confidence masks and trigger refinement; 
and (3) developing iterative refinement mechanisms that allow the model to 
correct initial segmentation errors through multi-step reasoning.

\end{document}